\documentclass[10pt,twocolumn,letterpaper]{article}
\pdfoutput=1
\usepackage[pagenumbers]{iccv} 

\definecolor{iccvblue}{rgb}{0.21,0.49,0.74}
\usepackage[pagebackref,breaklinks,colorlinks,linkcolor=iccvblue,citecolor=iccvblue,urlcolor=magenta]{hyperref}
\usepackage{verbatim}
\usepackage{pifont}
\usepackage{bbding}
\usepackage{multirow}
\usepackage{url}

\usepackage{mathrsfs}
\usepackage{float}
\def\paperID{11315} 
\def\confName{ICCV}
\def\confYear{2025}

\title{Video Individual Counting for Moving Drones  }

\author{Yaowu Fan\textsuperscript{1} \quad  Jia Wan\textsuperscript{2} \quad Tao Han\textsuperscript{3} \quad Antoni B. Chan\textsuperscript{4} \quad Andy J. Ma\textsuperscript{1 * \Envelope}\\
\textsuperscript{1}Sun Yat-sen University \quad \textsuperscript{2}Harbin Institute of Technology (Shenzhen)\\
\textsuperscript{3}Hong Kong University of Science and Technology  \quad \textsuperscript{4}City University of Hong Kong \\
{\tt\small \{fywyukee, jiawan1998, hantao10200\}@gmail.com, abchan@cityu.edu.hk, majh8@mail.sysu.edu.cn}
}

\begin{document}
\maketitle
\renewcommand{\thefootnote}{}
\footnotetext{* A.J. Ma is also with the Guangdong Province Key Laboratory of Information Security Technology, China, and the Key Laboratory of Machine Intelligence and Advanced Computing, Ministry of Education, China.}
\footnotetext{\Envelope \hspace{2pt} Corresponding author.}
\renewcommand{\thefootnote}{\arabic{footnote}} 
\begin{abstract}
Video Individual Counting (VIC) has received increasing attention for its importance in intelligent video surveillance.
Existing works are limited in two aspects, i.e., \textbf{dataset} and \textbf{method}.
Previous datasets are captured with fixed or rarely moving cameras with relatively sparse individuals, restricting evaluation for a highly varying view and time in crowded scenes.
Existing methods rely on localization followed by association or classification, which struggle under dense and dynamic conditions due to inaccurate localization of small targets.
To address these issues, we introduce the  MovingDroneCrowd Dataset, featuring videos captured by fast-moving drones in crowded scenes under diverse illuminations, shooting heights and angles.  We further propose a \textbf{S}hared \textbf{D}ensity map-guided \textbf{Net}work (\textbf{SDNet}) using a Depth-wise Cross-Frame Attention (\textbf{DCFA}) module to directly estimate shared density maps between consecutive frames, from which the inflow and outflow density maps are derived by subtracting the shared density maps from the global density maps.
The inflow density maps across frames are summed up to obtain the number of unique pedestrians in a video. 
Experiments on our datasets and publicly available ones show the superiority of our method over the state of the arts in highly dynamic and complex crowded scenes.
Our dataset and codes have been released publicly\footnote{\url{https://github.com/fyw1999/MovingDroneCrowd}}.
\vspace{-0.3cm}
\end{abstract}    
\section{Introduction}
\label{sec:intro}

Crowd counting is a fundamental task in crowd analysis to estimate the pedestrian density and quantity in images or videos. 
This task plays an important role in safety monitoring and early warning of stampedes to prevent crowd disasters caused by abnormal congestion \cite{Crowd_analysis_survey}.
\begin{figure}[t]
	\centering
	\includegraphics[width=1\linewidth]{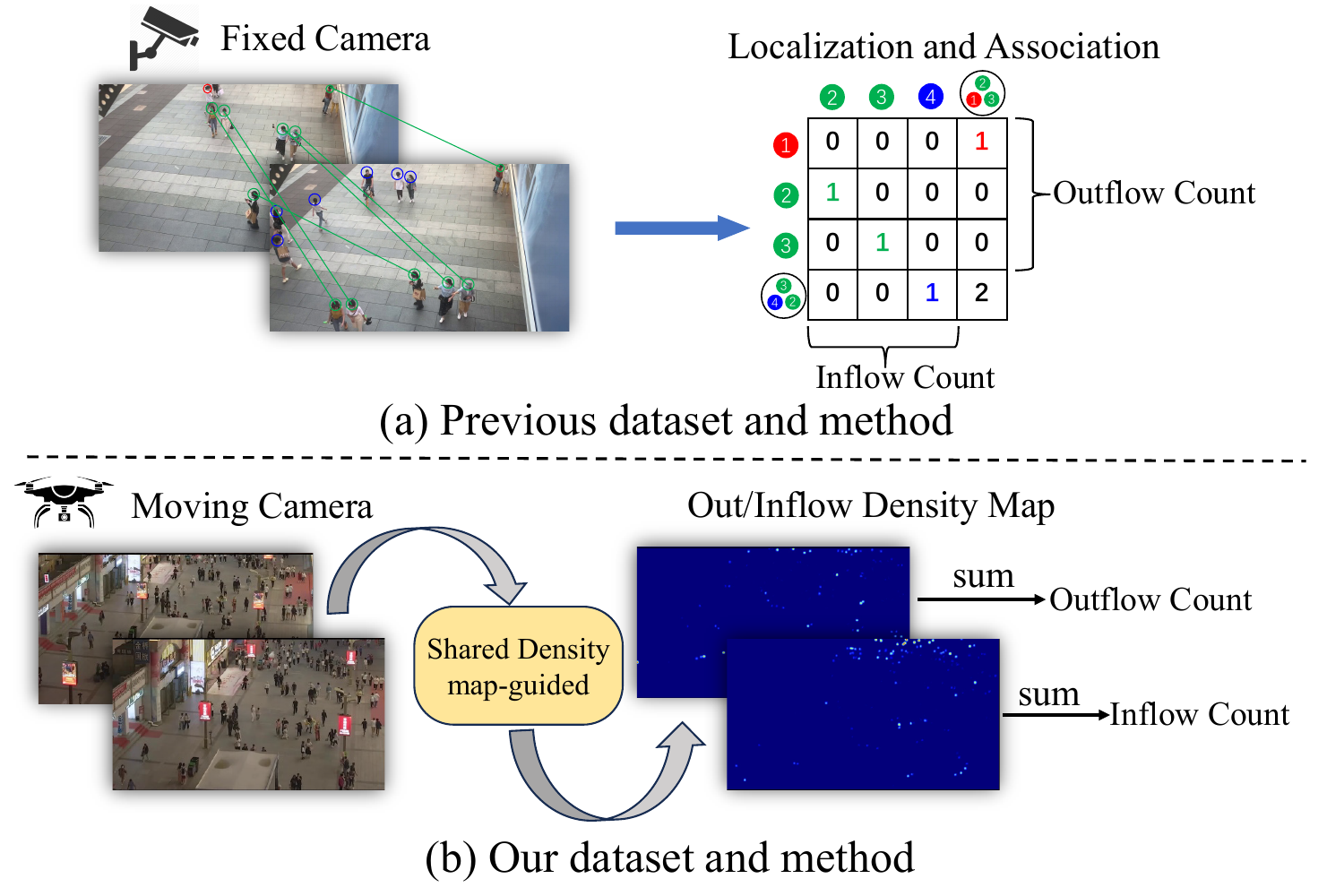}
    \vspace{-0.7cm}
	\caption{Comparison between our dataset/method and existing ones. \textbf{Dataset:} Existing datasets are captured by fixed or hardly moving cameras with sparse targets, while our data is collected from high-speed moving drones in crowded scenes.
    \textbf{Method:} 
    Existing methods first localize pedestrians and then perform cross-frame association or classification. 
    They fail on challenging datasets like ours due to the difficulty in accurately localizing pedestrians under crowded and complex scenes.
    Instead, our shared density map-guided method adopts a more learnable and optimizable approach by first estimating shared density maps via cross-frame attention and then inferring inflow and outflow density maps, leading to better performance under challenging scenarios.
    }
	\label{fig:comparison_dataset_method}
    \vspace{-0.6cm}
\end{figure}

Previous works primarily focus on crowd counting in images from handheld cameras, smartphones, and fixed surveillance cameras \cite{csrnet, Zhang_2016_CVPR_MCNN, Liu_2019_CVPR_Context_Aware, Attention_Scaling, P2P, RSMPL, LCSD}. 
While achieving remarkable progress, these methods are gradually failing to meet the demands of complex and dynamic real-world scenarios. On the one hand, these images are often captured at low heights and cover only limited regions. As a result, the perspective effect causes heads in regions that are far away from the cameras occlude each other, leading to inaccuracies in counting. 
On the other hand, counting in images provides only the number of pedestrians in a specific location at a given moment.
It fails to meet the real-world needs for estimating the number and density of pedestrians over large areas and periods of time, such as in pedestrian streets or crowded squares. 

To address the issues caused by ground-based cameras, existing works \cite{bahmanyar2019mrcnet, VisDrone-CC2021, RGB-T_CCD, VisDroneDatasets, DTCMDC} collect a series of drone-based datasets. Nevertheless, most of them are image-level or captured from a fixed drone viewpoint, restricting the monitoring of crowdedness within a limited view and time.
Although a drone video dataset is introduced in \cite{uavvic}, it includes both vehicles and pedestrians, resulting in a relatively low pedestrian density. 
Moreover, since their videos were collected by drones in suburbs with uniform shooting heights, angles, and lighting, they may not be able to represent complex and crowded real-world scenes.

Besides dataset limitations, accurately counting pedestrians with different identities in a video (a.k.a. video individual counting~\cite{DRNet}) remains challenging. The most straightforward idea is to apply multi-object tracking (MOT) techniques~\cite{TrackFormer, TRMOT, ByteTrack, MeMOT, DanceTrack} and count the tracklets. Since MOT-based methods are typically designed for sparse scenes with large targets, they fail in crowded scenes with low-resolution targets.
Recently, several methods \cite{DRNet, uavvic, PGDTR} have been proposed specifically for this task, which localize persons in each frame and then associate or classify them between two consecutive frames to infer inflow count.
Despite these efforts, all methods heavily depend on accurate pedestrian localization, which is unreliable in dense crowds.
Poor localization leads to degraded association or classification, resulting in significant counting deviations across videos. 
Hence, the localization-then-association or localization-then-classification paradigm is fragile in complex environments with dense crowds, particularly when captured by a fast-moving drone. The most related method to ours is \cite{FMDC}, which directly predicts inflow and outflow masks and then multiplies them with global density maps to obtain inflow and outflow density maps. However, we argue that directly predicting frame-specific pedestrians from two frames is more difficult. In contrast, our method first estimates the shared density maps between frames and then infers the inflow and outflow density maps.

The dataset and method limitations in existing works are illustrated in Fig \ref{fig:comparison_dataset_method}. 
To overcome these limitations, we collect a MovingDroneCrowd Dataset and propose a shared density map-guided method for video individual counting.
Unlike existing datasets, our dataset specifically focuses on crowded scenes captured by moving drones under diverse and complex conditions, including pedestrian streets, tourist attractions, and squares. It features complex camera motion patterns and a wider variety of light conditions, shooting angles, and shooting heights, making the task of video individual counting highly challenging and existing methods less effective. 
For methodology, the proposed method is inspired by the observation in image-level crowd counting that density map-based methods yield lower counting errors than localization-based ones in crowded scenes, and by the intuition that identifying shared objects between two sets is easier and more learnable than detecting set-specific ones.

Specifically, we design a Depth-wise Cross-Frame Attention (DCFA) module to learn the respective shared density maps for two adjacent frames, where each shared density map includes the density of pedestrians that appear in both the current and the adjacent frame.
The proposed DCFA takes multi-scale features from two consecutive frames as input and computes cross-frame attention across features with different scales. 
The features of each frame output by the DCFA module are decoded by the shared density map decoder to obtain their respective shared density maps.
Finally, outflow and inflow density maps are estimated by subtracting the shared density maps from the global density maps. 
During testing, unique pedestrians in a video clip are counted  by summing the inflow density maps across frames.
Our method is weakly supervised, which requires only inflow and outflow labels indicating whether pedestrians enter or exit the view.
The contributions of this paper are summarized as follows:

\begin{itemize}[labelindent=0em] 
 
\item[$\bullet$] We collect a video-level individual counting dataset captured by fast-moving drones in various crowded scenes. Compared to prior datasets, our one is with higher crowd density, more complex camera motions, and greater variations in lighting, shooting angles and heights.

\item[$\bullet$] We propose a shared density map-guided VIC method that bypasses the challenging localization step and instead adopts a more learnable manner by first learning shared pedestrian density maps between consecutive frames.
\item[$\bullet$] We design a Depth-wise Cross-Frame Attention (DCFA) module to extract shared density maps, 
which are then subtracted from the global density maps to obtain accurate inflow density.
\item[$\bullet$] Experiments on our dataset and publicly available ones show that the proposed method outperforms the state of the arts in highly dynamic, dense, and complex scenes.
\end{itemize}

\section{Related Works}
\subsection{Image-level crowd counting} 
In early works of crowd counting \cite{Bayesian_Poisson, MSMS, OCLSI}, handcrafted features were utilized to regress the number of persons in images. Spatial information is leveraged to improve performance in  \cite{LTCO} by learning a mapping between image features and density maps. Nowadays, CNNs or Transformers are used to map the image features to density maps. These works tackle challenges such as perspective effects \cite{Reverse_Perspective, Reverse_Perspective2, Perspective-Guided}, domain differences \cite{liu2022leveraging, Domain-General_Du_Deng_Shi_2023, Striking_a_Balance, HQITDR, NLT, FSCCDCU}, or scale variations \cite{Redesigning_Multi-Scale, STNet, STEERER}. Though density map-based methods can provide more accurate counts, they cannot determine the exact coordinates of individuals, especially in regions far away from the camera. To this end, crowd localization is proposed to directly regress the coordinates of each person using neural networks \cite{P2P, E2ETCL}.
\cite{CLRNet, VSCrowd, LCSTNVCC} leverage adjacent frames to enhance counting and localization performance in the target frame. They still count the same person multiple times across different frames, so they are still categorized as image-level crowd counting. 
Traditional image-level methods can only perform counting within a fixed region at a single time point, whereas our method enables counting over dynamically changing views.
\subsection{Video-level crowd counting}
Counting pedestrians with different identities over a period of time is more meaningful. We classify this task as video-level crowd counting, and in work \cite{DRNet}, it is also defined as Video Individual Counting. Intuitively, MOT techniques \cite{Headhunter-T, ByteTrack, BoT_SORT} offer a potential solution. However, these methods struggle in highly crowded scenes with several occlusions and are ineffective in handling rapid camera movements. 
Han \textit{et al.} \cite{DRNet} decomposes this task as a pedestrian association problem between two consecutive frames. 
Liu \textit{et al.} \cite{uavvic} further proposed a weakly-supervised group-level matching method. 
\cite{PGDTR} regress the coordinates of person and then classify them into shared, inflow, and outflow person. However, these methods require localizing individuals in each frame, followed by association or classification, where localization errors can severely affect accuracy. Wan \textit{et al.} \cite{FMDC} proposed a density map-based method that predicts inflow and outflow masks and then multiplies the masks with global density maps to obtain inflow and outflow density maps, but this process is difficult to learn and optimize.
In contrast, our method formulates this task in a more learnable manner by first estimating the shared density maps and then inferring inflow and outflow density maps. 

\subsection{Drone-based crowd counting datasets}
Currently, datasets for crowd counting from a drone perspective remain relatively scarce. Bahmanyar \textit{et al.}\cite{bahmanyar2019mrcnet} collected an aerial crowd dataset using DSLR cameras mounted on a helicopter. The datasets proposed in \cite{VisDrone-CC2021, RGB-T_CCD} are formed in RGB and thermal pairs captured by drones. However, these datasets are all image-level, meaning they only allow counting the number of persons at a specific moment within a fixed view. The multi-object tracking dataset \cite{VisDroneDatasets} for drone perspectives contains video clips with dense crowds. However, during annotation, these crowded regions were entirely ignored.
Luo \textit{et al.} \cite{DBJDLT, DTCMDC} released a video-level drone crowd dataset, but the video clips were captured by hovering drones, with each clip covering only a fixed field of view, similar to image-level datasets. The dataset UAVVIC \cite{uavvic} collects video clips captured by drones in relatively simple and uniform conditions. It includes not only pedestrians but also a large number of vehicles, leading to a lower pedestrian density. Compared to them, our dataset is captured by fast-moving drones under more complex conditions, including denser crowds, more challenging lighting, and more diverse flying altitudes and camera angles.
 
\begin{figure}[t]
	\centering
	\includegraphics[width=1\linewidth]{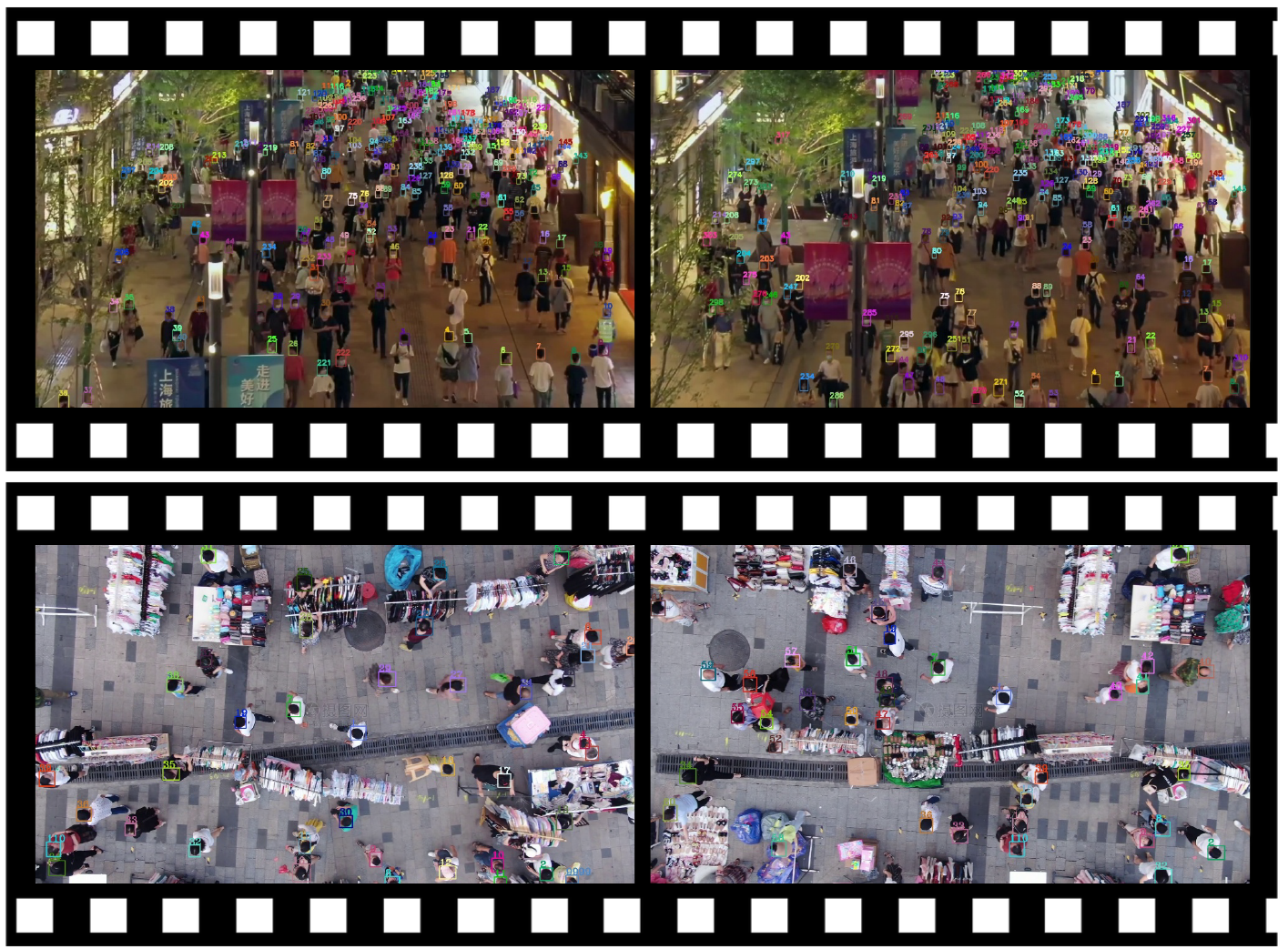}
	\vspace{-0.7cm}
	\caption{Two example clips from our dataset. The head bounding boxes and ID annotations are presented in each frame. The diverse light conditions, shooting angles, heights and densely packed pedestrians make it a highly challenging dataset. Only two frames per clip are shown to save space and provide a clearer presentation. Zoom in to see more details.}
	\label{fig:dataset_examples}
    \vspace{-0.5cm}
\end{figure}
\section{MovingDroneCrowd Dataset}
\label{chap:MovingDroneCrowd}
To promote practical crowd counting, we introduce MovingDroneCrowd — a video-level dataset specifically designed for dense pedestrian scenes captured by moving drones under complex conditions. Notably, our dataset provides precise bounding box and ID labels for each person across frames, making it suitable for multiple pedestrian tracking from drone perspective in complex scenarios. 
We detail the dataset and compare it with existing ones below. 

\textbf{Data Processing and Scale}: Due to strict regulations on drone flights, we obtained raw drone videos from the internet using keywords like ``aerial'', ``drone'', ``pedestrian flow'',  and  ``pedestrian street''. The raw videos were first segmented into clips covering entire locations. To reduce redundancy, each clip was downsampled to 1fps, 3fps, or 6fps based on drone speed. Some drone videos have very narrow shooting angles, making pedestrians farther from the camera appear extremely blurry. To alleviate the difficulty of annotation, these clips are cropped until the pedestrians within the shooting range can be identified by annotators. Finally, 89 clips (4940 frames) with resolutions of 720p, 1080p, 2K, and 4K are obtained.

\textbf{Annotation}: The annotation process was carried out by 10 well-trained annotators using the labeling tool DarkLabel \footnote{\url{https://github.com/darkpgmr/DarkLabel}}  and took a month to complete. Each annotator was asked to label bounding boxes that tightly enclose pedestrians' heads and assign unique IDs to different individuals in an entire video. Once the annotations were completed, the clips were reassigned to different annotators for error checking and revision. Finally, \textbf{325542} head bounding boxes and \textbf{16154} tracklets were obtained. Fig. \ref{fig:dataset_examples} displays two video clips from our dataset, with head bounding boxes and ID labels, illustrating their diverse lighting conditions, shooting angles, and heights, as well as higher crowd density. These attributes make our dataset more challenging and distinguish it from previous datasets. 

\textbf{Dataset Partition}: The dataset is split into training (70\%), testing (20\%), and validation (10\%) sets at the \textbf{scene level}, ensuring no overlapping scenes. This setup places higher demands on the algorithm's generalization ability. In addition, the data split process ensures that each set contains diverse data.

\begin{figure}[t]
	\centering
	\includegraphics[width=0.8\linewidth]{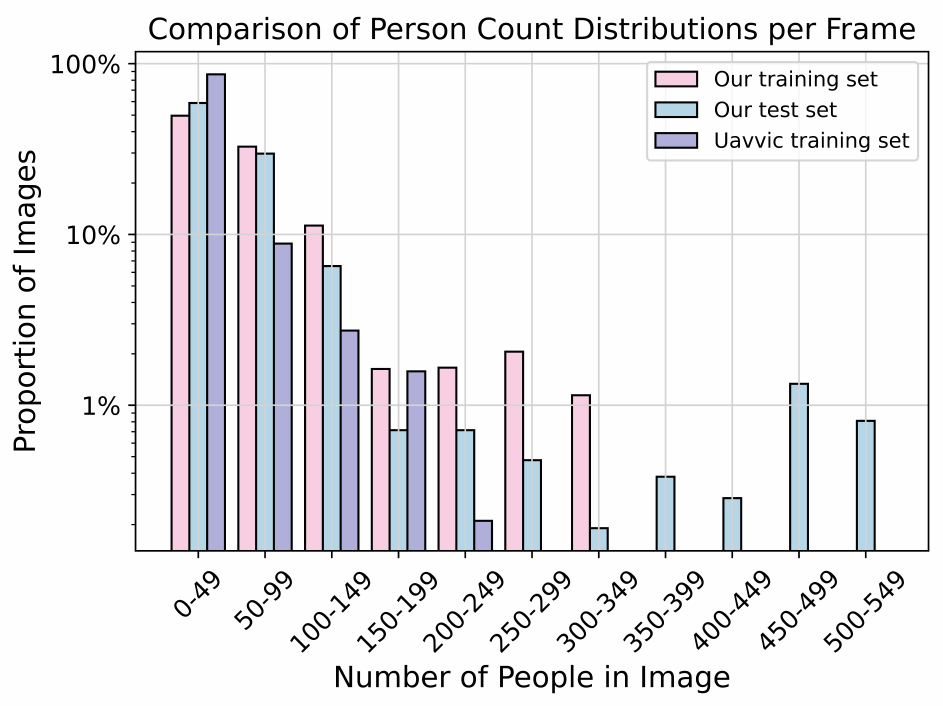}
	\vspace{-0.4cm}
	\caption{Comparison of pedestrian count distribution per frame between our dataset and UAVVIC.}
	\label{fig:num_distribution_compare}
    \vspace{-0.5cm}
\end{figure}

\textbf{Comparison}: As shown in Table \ref{tab:dataset_comparison}, we compare our dataset against recent video datasets. Compared with the previous drone dataset \cite{uavvic}, ours specifically focuses on dense pedestrians and has diverse light conditions, shooting angles, and shooting heights, as well as more complex motion patterns. Fig. \ref{fig:num_distribution_compare} shows the pedestrian count distribution per frame of moving data between our dataset and UAVVIC. Because UAVVIC's test set is unavailable, we only include the comparative results of the training set. Based on the statistical results, most moving frames in UAVVIC contain fewer than 50 pedestrians, whereas our dataset exhibits a higher proportion of frames in the ranges of $50 - 99$ and $100-149$, which correspond to typical crowded scenarios. Additionally, our training set has frames distributed in the more crowded range of $250-349$, and our test set includes some extremely crowded moving frames with pedestrian count in the range of $350-549$, whereas UAVVIC lacks. In summary, our dataset offers a more diverse and challenging pedestrian count distribution.
 

\begin{table}[t]
\tabcolsep=0.01cm
\resizebox{\linewidth}{!}{
\begin{tabular}{cccccccccc}
\hline
Dataset          & Perspective  & Moving & MFR & MPR &MPPF& Light& Height & Angle & IDs \\ \hline
CroHD            & Surveillance & \ding{55}       &  0           & 0   &0         &day\&night& Fixed                 &  Fixed                      &  \checkmark    \\
VSCrowd          & Surveillance & \ding{55}       &  0           & 0    &0        &day\&night&  Fixed                &    Fixed                    & \checkmark     \\ \hline
DroneCrowd       & Drone        & \ding{55}       &  0          & 0    &0        &day\&night&  Fixed                &  Fixed                      & \checkmark     \\
UAVVIC           & Drone        &  \checkmark\kern-1.1ex\raisebox{.7ex}{\rotatebox[origin=c]{125}{--}}      & 51\%            & 39\%            &32 &day & $\sim$ 20m                 &  $\sim$ 90°                      & \ding{55}     \\
MovingDroneCrowd & Drone        &  \checkmark      & 100\%            & 100\%             &66 &day\&night& $\sim$ 3-20m                 & $\sim$ 45-90°                      & \checkmark     \\ \hline
\end{tabular}
}
\vspace{-0.2cm}
\caption{Comparison of recent video datasets. MFR represents the proportion of moving frames to all frames, MPR denots the proportion of pedestrians in moving frames to the total number of pedestrians, and MPPF is the average number of pedestrians per frame in moving frames. Our dataset is captured in highly dynamic and complex scenarios, making it the most challenging.}
\label{tab:dataset_comparison}
\vspace{-0.5cm}
\end{table}

\section{Methodology}
\subsection{Problem Formulation}
Formally, the training set $\mathcal{V}_t = \{ \textbf{V}_i, \textbf{L}_i\}^{N_t}_{i=1}$ consists of $N_t$ video clips and annotations, where the $i^\text{th}$ video $\textbf{V}_i=\{V_j\}_{j=1}^{n_i}$ has $n_i$ frames, and $\textbf{L}_i=\{ P_j, ID_j\}_{j=1}^{n_i}$ provides the coordinates and identities of the person in each frame of video $\textbf{V}_i$. Notably, our method is weakly supervised and does not require ID labels, making it applicable even when only inflow $I_j$ and outflow labels $O_j$ that indicate pedestrian entries and exits are provided.

For consecutive frames $V_j$ and $V_{j+\delta}$ (with a fixed interval $\delta$), our method estimates the outflow density map $\mathbf{\hat{D}}^{out}_j$ for $V_j$ and inflow density map $\mathbf{\hat{D}}^{in}_{j+\delta}$ for $V_{j+\delta}$. The sum of $\mathbf{\hat{D}}^{out}_j$  gives the number of pedestrians in $V_j$ who exit the view of $V_{j+\delta}$, while the sum of $\mathbf{\hat{D}}^{in}_{j+\delta}$ represents the number of pedestrians entering the view of $V_{j+\delta}$. Consequently, the total number of unique pedestrians in video $\mathbf{V}_i$ can be computed as:
\begin{equation}
\setlength\abovedisplayskip{2pt}
\setlength\belowdisplayskip{2pt}
    M(\mathbf{V}_i) \approx M(V_1) + \sum^{(n_i/\delta)-1}_{k=1} \mathrm{sum}(\mathbf{\hat{D}}^{in}_{1+k\times\delta}),
\end{equation}
where $M(V_1)$ represents the number of persons in the first frame, and $\mathbf{\hat{D}}^{in}_{1+k\times\delta}$ is the inflow density map of frame $V_{1+k\times\delta}$ relative to frame $V_{1+(k-1)\times\delta}$.

\subsection{Overall Framework}
\begin{figure*}[t]
	\centering
	\includegraphics[width=0.8\linewidth]{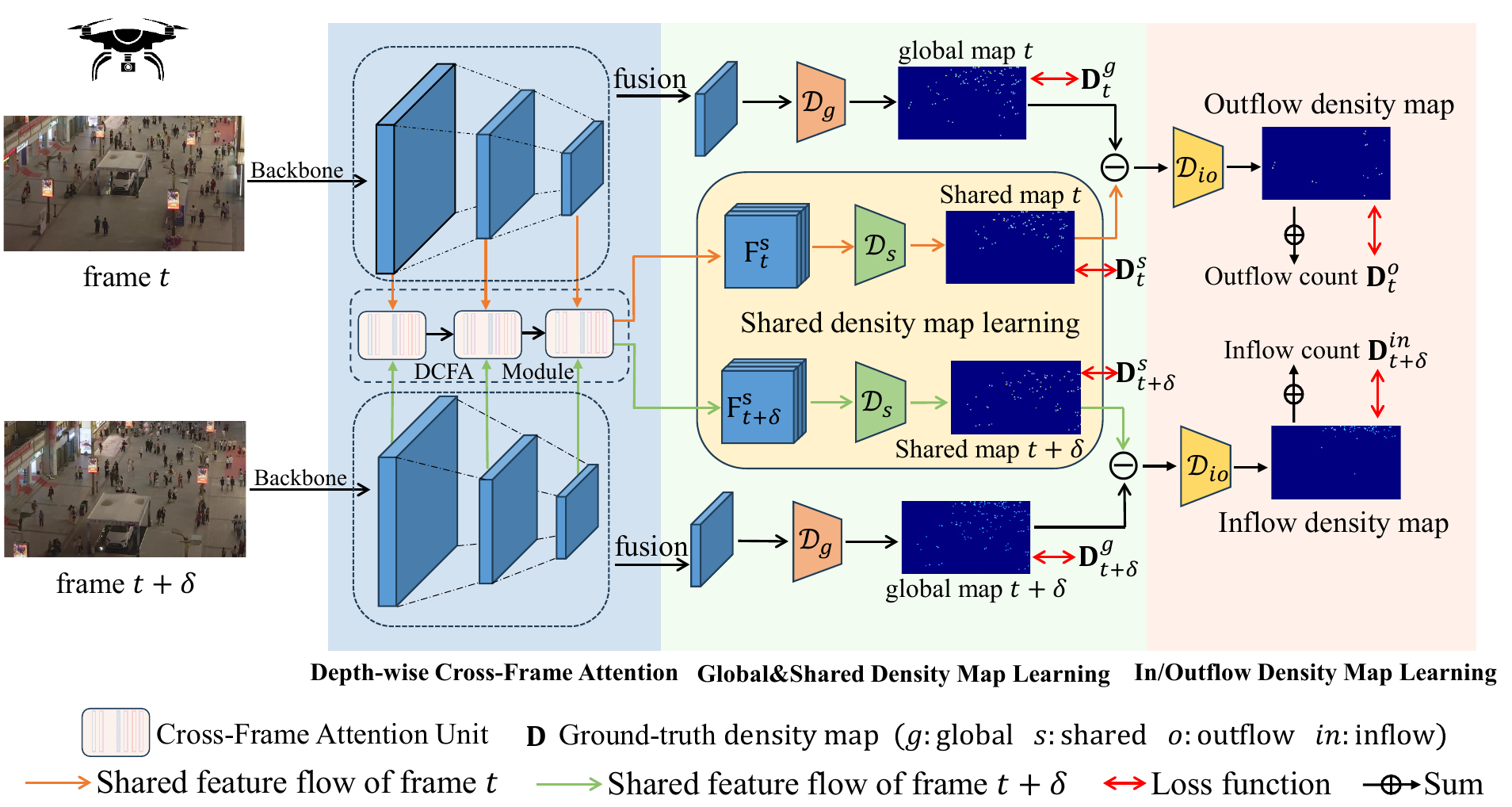}
	\vspace{-0.3cm}
    	\caption{The pipeline of our shared density map-guided VIC method. First, multi-scale features are extracted using a shared-weight CNN and FPN. The DCFA module computes cross-frame attention across features at all scales to obtain the shared features, while global features are obtained by fusing the multi-scale features. Then, a global decoder and a shared decoder generate global and shared density maps for each frame. Finally, the inflow-outflow decoder processes the difference between global and shared density maps to produce the outflow density map for the first frame and the inflow density map for the second frame. During testing,  simply accumulating the sum of the inflow density maps across all frames yields the total number of unique pedestrians in the entire video. }
	\label{fig:pipeline}
    \vspace{-0.5cm}
\end{figure*}

To achieve the goal mentioned above, \textit{i.e.}, estimating the inflow density map for each frame, we first estimate the shared density map, as illustrated in Fig. \ref{fig:pipeline}. Specifically, given two consecutive frames $V_j$ and $V_{j+\delta}$, we first extract their multi-scale features $\mathcal{F}_j$ and $\mathcal{F}_{j+\delta}$. Then, the extracted multi-scale features pass through our proposed Depth-wise Cross-Frame Attention module to obtain shared features $\mathbf{F}^s_j$ and $\mathbf{F}^s_{j+\delta}$ for each frame. The shared density map decoder  $\mathcal{D}_s$ maps the shared features to shared density maps  $\mathbf{\hat{D}}^{s}_j$ and  $\mathbf{\hat{D}}^{s}_{j+\delta}$. Meanwhile, the multi-scale features of each frame are fused and then mapped to global density maps $\mathbf{\hat{D}}_j^g$ and $\mathbf{\hat{D}}_{j+\delta}^g$ through the global density map decoder $\mathcal{D}_g$. Finally, the differences between the global and shared density maps are used to derive the outflow density map $\mathbf{\hat{D}}^{out}_j$ for $V_j$ and inflow density map $\mathbf{\hat{D}}^{in}_{j+\delta}$ for $V_{j+\delta}$.

\subsection{Depth-wise Cross-frame Attention}
To learn the shared and global features, we first extract multi-scale features. Given sampled consecutive frames $V_j$ and $V_{j+\delta}$, a shared-weight backbone network and a Feature Pyramid Network extract multi-scale features $\mathcal{F}_j$ and $\mathcal{F}_{j+\delta}$, where $\mathcal{F}_j=\{\mathbf{F}_j^i\}_{i=1}^{N_f}$, and $N_f$ is the number of multi-scale feature levels. The dimension of the $i$-th scale feature $\mathbf{F}_j^i$ is $C \times H/2^{(i+1)} \times W/2^{(i+1)}$. Here, $H$ and $W$ are the height and width of input image, respectively, and $C$ is the number of feature channels.

\begin{figure}[t]
	\centering
	\includegraphics[width=0.8\linewidth]{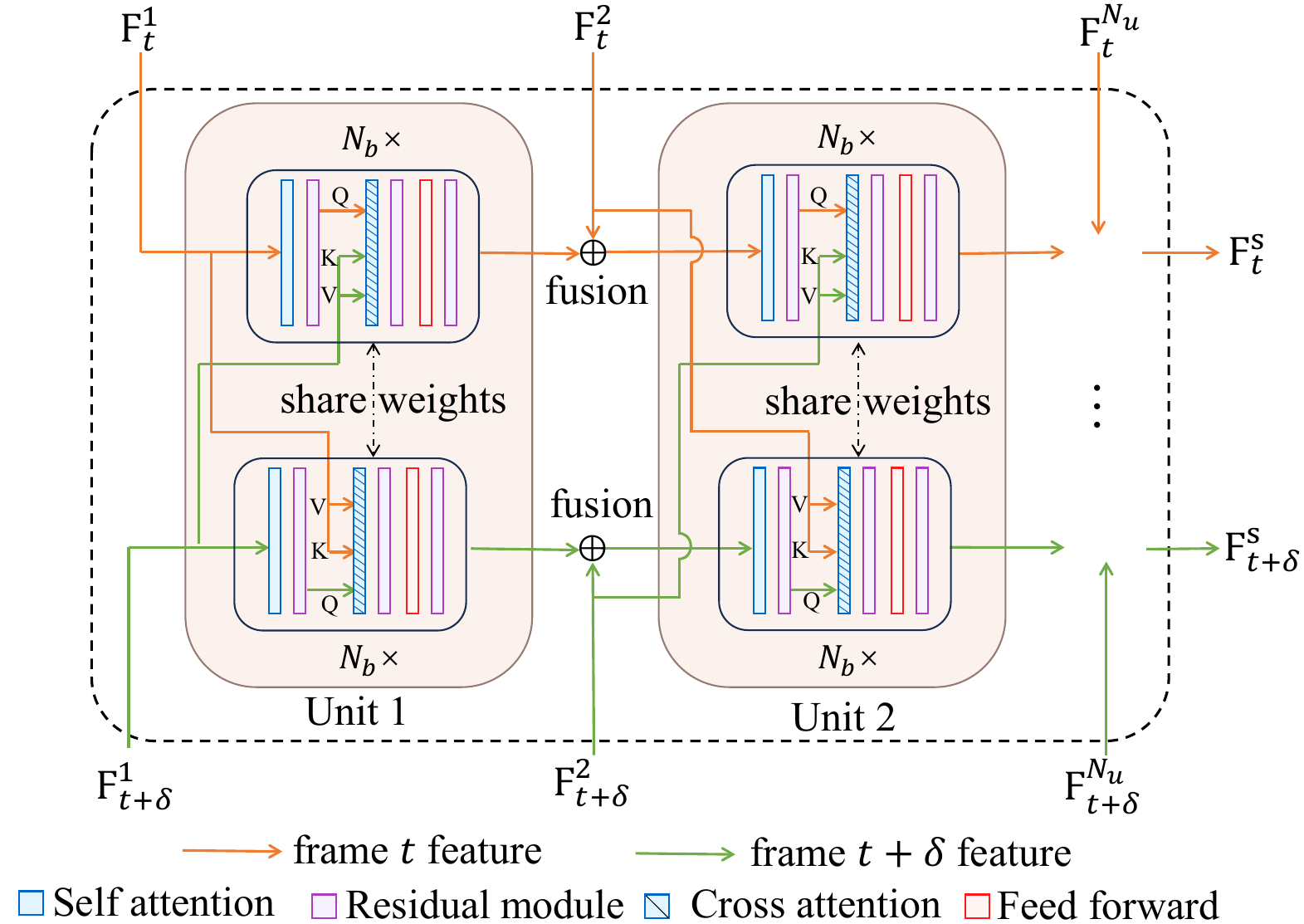}
	\vspace{-0.2cm}
	\caption{The details of our proposed DCFA module. It contains $N_u$ cross-frame attention units, each comprising $N_b$ cross-frame blocks. The number of units matches the multi-scale feature levels. For the $i^\text{th}$ unit, cross-frame attention is computed using the fused feature of the first frame's feature at $i^\text{th}$ scale and the output of the $(i-1)^\text{th}$ unit as the \textit{query} and the second frame's feature at $i^\text{th}$ scale level as \textit{key} and \textit{value}. The final unit's output is the shared feature of the first frame. Swapping the roles of the two frames, yields the shared feature of the second frame.} 
	\label{fig:DCFA}
    \vspace{-0.5cm}
\end{figure}

With the extracted multi-scale features, our designed Depth-wise Cross-Frame Attention (DCFA) module is used to learn shared features for each frame. The details of our DCFA are illustrated in Fig. \ref{fig:DCFA}. DCFA consists of $N_u$ cross-frame attention units, each containing $N_b$ cross-frame attention blocks. The number of units in DCFA corresponds to the number of scale levels in the multi-scale features. 
When computing the shared feature of frame $V_j$, the first cross-frame attention unit directly takes $\mathbf{F}_j^1$ as input, while  
for $i^{\text{th}}$ unit ($i > 1$), the $i^{\text{th}}$ scale feature $\mathbf{F}_j^i$ of frame $V_j$ is first fused with the output $\hat{\mathbf{F}}^{i-1}_j$ of the $(i-1)^{\text{th}}$ unit:
\begin{equation}
\setlength\abovedisplayskip{3pt}
\setlength\belowdisplayskip{3pt}
   \tilde{\mathbf{F}}^{i}_j= \mathrm{Fusion}(\hat{\mathbf{F}}^{i-1}_j, \mathbf{F}^{i}_j).
\end{equation}
The process of computing the output of the $i^{\text{th}}$ unit is then performed as follows:
\begin{equation}
\setlength\abovedisplayskip{3pt}
\setlength\belowdisplayskip{3pt}
\resizebox{0.6\hsize}{!}{$
\begin{aligned}
    {\mathbf{F}^i_j}' &= \mathrm{MSA}(\mathrm{LN}(\tilde{\mathbf{F}}^{i}_j)) + \tilde{\mathbf{F}}^{i}_j, \\
    {\mathbf{F}^i_j}'' &= \mathrm{MCA}(\mathrm{LN}({\mathbf{F}^i_j}'), \mathbf{F}^i_{j+\delta}) + {\mathbf{F}^i_j}', \\
    \hat{\mathbf{F}}^{i}_j &= \mathrm{MLP}(\mathrm{LN}({\mathbf{F}^i_j}'')) + {\mathbf{F}^i_j}'', 
\end{aligned}
$}
\label{eq:cross_frame_attention_block}
\end{equation}
where  $\mathrm{LN}$ denotes layer normalization, $\mathrm{MSA}$ represents multi-head self-attention layer, and $\mathrm{MCA}$ refers to multi-head cross-attention layer. The computation of the $\mathrm{MCA}$ layer in Eq. \ref{eq:cross_frame_attention_block} indicates that the multi-scale features from frames $V_j$ and $V_{j+\delta}$ are set as the \textit{query} and \textit{key}, respectively. This process can be formulated as follows :
\begin{equation}
\setlength\abovedisplayskip{3pt}
\setlength\belowdisplayskip{3pt}
\resizebox{0.8\hsize}{!}{$
    \begin{aligned}
        Q_h={\mathbf{F}_j^i}'W^{Q}_h, \hspace{0.2cm} &K_h=\mathbf{F}_{j+\delta}^iW^{K}_h, \hspace{0.2cm} V_h=\mathbf{F}_{j+\delta}^iW^{V}_h, \\
        Head_h &= \mathrm{Softmax}(\frac{Q_hK_h^T}{\sqrt{D}})V_h,\\
        {\mathbf{F}^i_j}''= \hspace{0.1cm} &\mathrm{Concat}(Head_1,...,Head_H),
    \end{aligned}
    $}
\end{equation}
where $W^{Q}_h$, $W^{K}_h$ and $W^{V}_h$ are learnable projection matrices. Here, $h$ represents the $h^{\text{th}}$ dependent head, and the final output is obtained by concatenating the outputs of all heads. 


This process is repeated iteratively until the final cross-frame attention unit outputs $\hat{\mathbf{F}}^{N_u}_j$, serving as the shared feature $\mathbf{F}^s_j$ of $V_j$.
Similarly, swapping the roles of $\mathbf{F}_j^i$ and $\mathbf{F}_{j+\delta}^i$, i.e. setting  $\mathbf{F}_{j+\delta}^i$ as the \textit{query} and $\mathbf{F}_j^i$ as the \textit{key} and \textit{value}, yields the shared feature $\mathbf{F}^s_{j+\delta}$ for frame $V_{j+\delta}$. The DCFA module effectively integrates multi-scale features and captures rich cross-frame information, thereby learning features that retain only shared pedestrian information between the consecutive frames.

\subsection{Inflow/Outflow Density Map Learning}
To derive the inflow and outflow density maps, shared and global density maps for frames $V_j$ and $V_{j+\delta}$ are first decoded:
    \begin{equation}
    \mathbf{\hat{D}}^g_j = \mathcal{D}_g(\mathbf{F}_j^g), \quad \mathbf{\hat{D}}^s_j = \mathcal{D}_s(\mathbf{F}_j^s),
\end{equation}
where $\mathcal{D}_g$ and $\mathcal{D}_s$ denote global and shared density map decoders, respectively. They have identical architectures comprising of alternating convolutional layers and upsampling operations to progressively restore the resolution to match the input image size. Here, $\mathbf{F}_j^g$ is the global feature of $V_j$, obtained by directly fusing the multi-scale features in $\mathcal{F}_j$.

The global density maps contain the densities of all pedestrians in each frame, while the shared density maps only include densities for pedestrians appearing in both frames. Consequently, the outflow and inflow density maps can be obtained from the difference between the global and shared density maps:
\begin{equation}
\setlength\abovedisplayskip{3pt}
\setlength\belowdisplayskip{3pt}
\begin{aligned}
    \mathbf{\hat{D}}^o_j &= \mathcal{D}_{io}(\mathbf{\hat{D}}_j^g-\mathbf{\hat{D}}_j^s), \\
    \mathbf{\hat{D}}^{in}_{j+\delta} &= \mathcal{D}_{io}(\mathbf{\hat{D}}_{j+\delta}^g-\mathbf{\hat{D}}_{j+\delta}^s),
\end{aligned}
\end{equation}
where $\mathcal{D}_{io}$ is the inflow-outflow decoder that is composed of convolutional layers. Obviously, the outflow density map contains the density of pedestrians appearing only in frame $V_j$, while the inflow density map contains the density of those appearing only in frame $V_{j+\delta}$. During testing, summing the inflow density maps of all frames yields the total number of pedestrians in the video.

Our framework is trained with four MAE losses: global $\mathcal{L}_g$, shared $\mathcal{L}_s$, outflow $\mathcal{L}_o$, and inflow $\mathcal{L}_{in}$ density map loss. These losses are computed as follows:
\begin{equation}
\setlength\abovedisplayskip{3pt}
\setlength\belowdisplayskip{3pt}
\resizebox{0.9\hsize}{!}{$
\begin{aligned}
    \mathcal{L}_g &= \frac{1}{2N}\sum_{i=1}^{2N}|| \mathbf{\hat{D}}^g_i- \mathbf{D}^g_i||, \hspace{0.05cm} \mathcal{L}_s = \frac{1}{2N} \sum_{i=1}^{2N}|| \mathbf{\hat{D}}^s_i- \mathbf{D}^s_i||,\\
    \mathcal{L}_o &= \frac{1}{N} \sum_{i=1}^{N}|| \mathbf{\hat{D}}^o_{2i-1} - \mathbf{D}^o_{2i-1}||,  \hspace{0.05cm}  \mathcal{L}_{in} = \frac{1}{N}\sum_{i=1}^{N}|| \mathbf{\hat{D}}^{in}_{2i} - \mathbf{D}^{in}_{2i}||,
\end{aligned}$
}
\end{equation}
where $N$ is the number of image pairs in the training batch. $\mathbf{D}^g$, $\mathbf{D}^s$, $\mathbf{D}^o$, and $\mathbf{D}^{in}$ are ground-truth global, shared, outflow, and inflow density maps, respectively. Note that the ground-truth density maps can be generated using either fully supervised labels (IDs) or weakly supervised labels (inflow and outflow annotations). 

\begin{table*}[]
	\centering
        \resizebox{\textwidth}{!}{%
	\begin{tabular}{c|c|c|ccc|cc|cccc}
		\toprule
		\multirow{2}{*}{Method} &  \multirow{2}{*}{Venue} & \multirow{2}{*}{ID}& \multirow{2}{*}{MAE$\downarrow$} & \multirow{2}{*}{RMSE$\downarrow$} & \multirow{2}{*}{WRAE$\downarrow$} & \multirow{2}{*}{MIAE$\downarrow$} & \multirow{2}{*}{MIOE$\downarrow$} & \multicolumn{4}{c}{MAE on four different density levels} \\ 
        \cline{9-12} 
		 & &                      &                      &               &   &&     & D0        & D1        & D2        & D3      \\ 
         \midrule
	
		ByteTrack\cite{ByteTrack}           &  ECCV'22      &\ding{55}              &  153.17                     &  227.62                      & 63.82    & 13.25&11.22      & 83.38           & \underline{24.00}          & 325.00          &  441.33        \\
		BoT-SORT\cite{BoT_SORT}           &  arxiv'22        &  \checkmark        &   150.61                   & 223.46                     & 62.53         &  13.11 & 11.22&   82.46        &   \textbf{22.00}       & 327.00          & 430.00         \\ 
		OC-SORT\cite{OC-SORT}           &  CVPR'23        &  \ding{55}        &   203.56                  &  276.84                    & 87.75        & 10.90 &13.63&  101.46        &  232.00         & 405.00          & 569.33         \\ 
		DiffMOT\cite{DiffMOT}           &  CVPR'24          & \checkmark         & 229.17                      & 450.86                      & 71.27         & 23.01        &  21.41          &45.85          & 292.00&952.00& 761.67       \\ 
        \midrule
		DRNet\cite{DRNet}              &CVPR'22   &\checkmark  &  81.14                    & 126.34                     & 33.36                     & \underline{5.64}          &    \textbf{5.09 }     & 28.73          &   129.88  &  217.13 &  246.69           \\
		CGNet\cite{uavvic}              &CVPR'24   &\ding{55} &  \underline{66.06}                   & \underline{110.36}                     & \underline{29.16}                       & -          &   -        &  \underline{25.92}        &  111.00 &   144.00 &\underline{199.00}                 \\ 
        \midrule
		LOI\cite{LOI}              &ECCV'16    & \checkmark            &  241.77         &337.90                    &  99.63                     &-           & -         &110.13           & 294.46 & 467.57 &  719.33                  \\ 
        \midrule
            FMDC\cite{FMDC}              &WACV'24    & \ding{55}            &  120.31         &183.57                   &  48.82                     &8.21           & 6.40         &61.66           & 75.71 & \underline{54.92} &  411.09 \\
        \midrule
		\multirow{2}{*}{Ours}               &  \multirow{2}{*}{-}  & \multirow{2}{*}{\ding{55}} & \textbf{41.00}                     & \textbf{58.34}                      &  \textbf{19.32}                  & \multirow{2}{*}{\textbf{5.50}}          & \multirow{2}{*}{\underline{6.39}}         & \multirow{2}{*}{\textbf{23.71}}        & \multirow{2}{*}{79.77} & \textbf{41.21} & \textbf{102.88}                  \\ 
        & & & \textcolor{red}{ $\downarrow$ 37.8\%}& \textcolor{red}{ $\downarrow$ 47.1\%}& \textcolor{red}{ $\downarrow$ 33.7\%} & & & &  & \textcolor{red}{ $\downarrow$ 24.9\%} & \textcolor{red}{ $\downarrow$ 48.3\%} \\
        \bottomrule
	\end{tabular}
        }
        \vspace{-0.2cm}
	\caption{Performance comparison on the MovingDroneCrowd dataset. D0 -- D3 respectively denote four pedestrian density ranges: [0, 150),   [150, 300), [300, 450), $\geq 450$. \textbf{Bold} indicates the best result, \underline{underline} denotes the second-best, and \textcolor{red}{red} shows the improvement of our method over the second-best. The performance advantage of our method becomes even more pronounced as crowd density increases.}
	\label{tab:comparison_on_our_dataset}
    \vspace{-0.2cm}
\end{table*}

\begin{table*}[]
\centering
\resizebox{\textwidth}{!}{%
\begin{tabular}{c|c|ccccc|cc|cc}
\toprule
\multirow{2}{*}{Method} & \multirow{2}{*}{Venue} & \multicolumn{5}{c|}{Overall} & \multicolumn{2}{c|}{Static} & \multicolumn{2}{c}{Dynamic} \\ \cline{3-7} \cline{8-9} \cline{10-11} 
                        &                        & MAE$\downarrow$   & RMSE$\downarrow$   & WRAE$\downarrow$  & MIAE$\downarrow$ & MOAE$\downarrow$  & MAE$\downarrow$          & RMSE$\downarrow$  & MSE$\downarrow$          & RMSE $\downarrow$       \\ 
                        \midrule
ByteTrack\cite{ByteTrack}                   & ECCV'22                        & \underline{14.19}      & \underline{21.51}      & \underline{68.92}      & \textbf{1.77}     & \textbf{2.09}  &9.40             &  10.21     & \underline{15.69}             & \underline{23.98}            \\
OC-SORT\cite{OC-SORT}                   &   CVPR'23                      &18.81       & 35.42      & 71.01      & 2.42     & 3.06  &7.20              & \underline{7.77}      & 22.44             &40.34             \\ \midrule
LOI \cite{LOI}                    & ECCV'16                       & 21.70      & 38.21      & 99.00      & -     &  -    & 11.12             &11.59             & 25.01             &43.29             \\ \midrule
CGNet\cite{uavvic}                   & CVPR'24                       & 24.95 & 52.57 & 83.82 & - & -    & \underline{6.80}         & 8.22         & 30.62        & 60.05       \\ \midrule
\multirow{2}{*}{Ours}                    &   \multirow{2}{*}{-}                     & \textbf{6.37}  & \textbf{11.01} & \textbf{46.01} & \multirow{2}{*}{\underline{1.81}} & \multirow{2}{*}{\underline{2.18}} & \textbf{3.30}         & \textbf{4.12}        & \textbf{7.33}        & \textbf{12.40}       \\
&&\textcolor{red}{ $\downarrow$ 55\%}& \textcolor{red}{ $\downarrow$ 48.8\%}& \textcolor{red}{ $\downarrow$ 33.2\%}&&&\textcolor{red}{ $\downarrow$ 51.5\%}&\textcolor{red}{ $\downarrow$ 47\%}&\textcolor{red}{ $\downarrow$ 53.3\%}&\textcolor{red}{ $\downarrow$ 48.3\%} \\
\bottomrule
\end{tabular}
}
\vspace{-0.2cm}
\caption{Performance comparison on validation set of UAVVIC. The results shows that our method consistently achieves the best results across overall, static, and dynamic scenes, demonstrating its effectiveness in both dynamic and sparse scenarios. }
\label{tab:comparison_on_uavvic}
\textbf{\vspace{-0.8cm}}
\end{table*}
\section{Experiments} 
Due to space limitations, please refer to the supplementary materials for more details on implementation details.

\subsection{Datasets}
Datasets UAVVIC and our MovingDroneCrowd are used for evaluation. A detailed description and comparison of these two datasets have been introduced above. 
\subsection{Evaluation Metrics}
Similar to image-level crowd counting, MAE and RMSE are used for evaluation, but they are computed at the video level.
Additionally, we also adopt the metric WRAE, MIAE, and MOAE defined in \cite{DRNet}. WRAE (Weighted Relative Absolute Errors) accounts for the impact of frame counts in different videos when computing relative errors. MIAE and MIOE measure the prediction quality of inflow and outflow, respectively. Please refer to \cite{DRNet} and its Supplementary for details. 

\subsection{Comparison with State of the Arts}

\textbf{Comparison Methods:} To demonstrate the superiority of our method, we compare it against a diverse range of related works. In addition to algorithms specifically designed for VIC, we also include other relevant approaches, such as multiple object tracking and cross-line crowd counting. 
\begin{figure*}[t]
	\centering
	\includegraphics[width=0.95\linewidth]{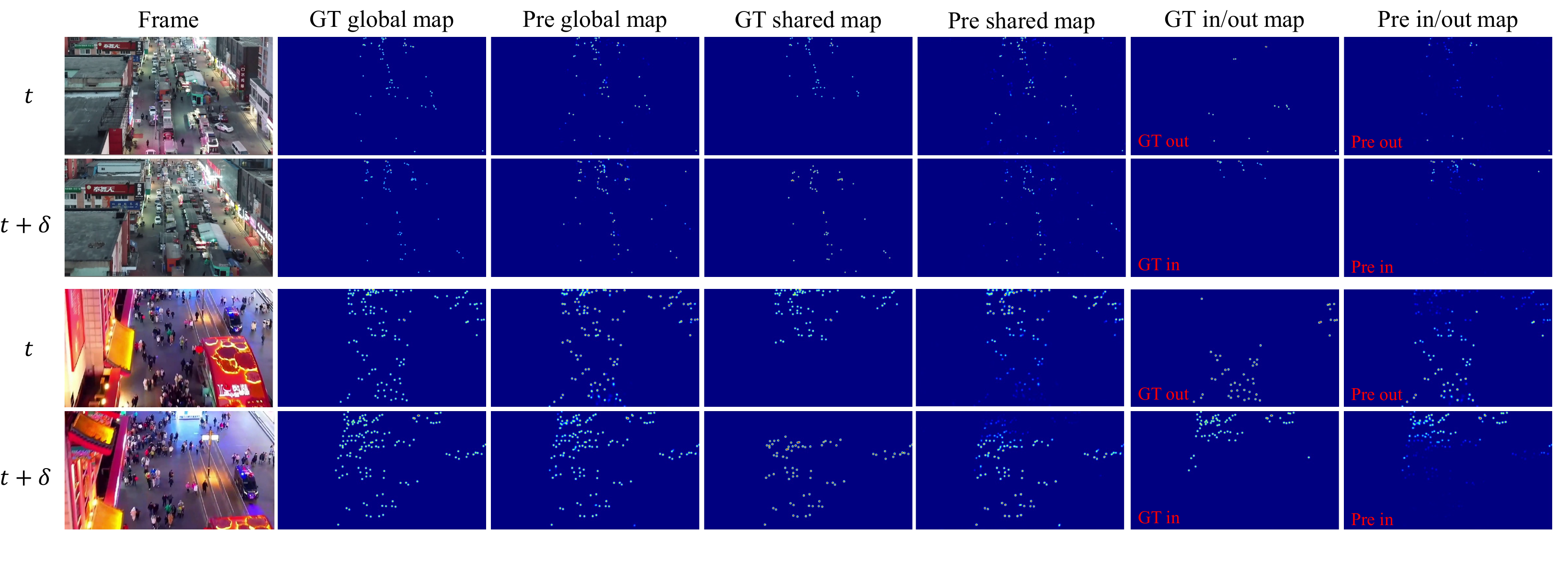}
	\vspace{-0.4cm}
	\caption{The visualization results of our method on MovingDroneCrowd. It presents the results of two consecutive frames. In addition to the global density map for each frame, the first frame includes its shared density map and outflow density map relative to the second frame, while the second frame includes its shared density map and inflow density map relative to the first frame.}
	\label{fig:visual}
    \vspace{-0.5cm}
\end{figure*}
\begin{table}[]
\tabcolsep=0.08cm
\resizebox{\linewidth}{!}{
\begin{tabular}{c|cc|ccccc}
\toprule
\multirow{2}{*}{Ablation}        & \multicolumn{2}{c|}{\multirow{2}{*}{Setting}} & \multirow{2}{*}{MAE$\downarrow$} & \multirow{2}{*}{RMSE$\downarrow$} & \multirow{2}{*}{WRAE$\downarrow$} & \multirow{2}{*}{MIAE$\downarrow$} & \multirow{2}{*}{MIOE$\downarrow$} \\
                                 & \multicolumn{2}{c|}{}                         &                      &                       &                       &                       &                       \\ \midrule
\multirow{4}{*}{Backbone}        & \multirow{2}{*}{VGG}         & w/o PE        & \textbf{41.00}       & \textbf{58.34}        & \textbf{19.32  }      & \textbf{5.50 }        & \textbf{6.39  }                    \\
                                 &                              & w/ PE         & 66.64                & 102.66                & 37.73                 & 7.80                  & 9.37                  \\ \cline{2-8} 
                                 & \multirow{2}{*}{ViT}         & w/o PE        & 98.56                & 142.83                & 48.50                 & 7.84                  & 8.07                  \\
                                 &                              & w/ PE         & 51.76                & 66.99                 & 24.90                 & 9.21                  & 8.10                  \\ \midrule
\multirow{2}{*}{Cross-Frame}     & \multicolumn{2}{c|}{DCFA}                     & \textbf{41.00}       & \textbf{58.34}        & \textbf{19.32  }      & \textbf{5.50 }        & 6.39                      \\
                                 & \multicolumn{2}{c|}{SCFA}                     & 70.42                & 90.13                 & 44.71                 & 5.80                  & \textbf{6.02}                  \\ \midrule
\multirow{2}{*}{Inflow Learning} & \multicolumn{2}{c|}{Direct}                   & 65.64                & 99.34                 & 47.12                 & 6.41                  & 7.63                      \\
                                 & \multicolumn{2}{c|}{SDNet}              & \textbf{41.00}       & \textbf{58.34}        & \textbf{19.32  }      & \textbf{5.50 }        & \textbf{6.39}                          \\ \bottomrule
\end{tabular}

}
\vspace{-0.2cm}
\caption{Ablation study for our method. ``Direct'' represents directly learning the inflow density map rather than first learning shared density map.  }
\label{tab:ablation_study}
\vspace{-0.6cm}
\end{table}

\noindent \textbf{Results on MovingDroneCrowd:} Table \ref{tab:comparison_on_our_dataset} compares our method with other approaches on our dataset MovingDroneCrowd. Our approach significantly outperforms others, reducing MAE and RMSE by $37\%$ and $47\%$, respectively, compared to the latest approach CGNet. For a more in-depth and detailed analysis, we divide the test scenes by pedestrian density and evaluate MAE under different density levels. As pedestrian density increases, other methods degrade sharply, while our method consistently maintains reasonable performance. The MOT-based methods completely fail in high-density scenes due to their reliance on individual detection and global identity association, which becomes infeasible in our dataset, including complex scenes with severe occlusion and rapid camera movements. VIC methods alleviate some issues but still rely on localization and cross-frame association, leading to unsatisfactory performance in highly crowded scenes. Density-based method FMDC \cite{FMDC} performs poorly despite avoiding localization and association, as directly predicting inflow and outflow masks is highly challenging. In contrast, our method first infers the more learnable shared density maps, and then derives the inflow and outflow maps, allowing it to achieve satisfying results even in complex and crowded scenes.
 
\noindent \textbf{Results on UAVVIC:} We also conduct comparative experiments on the drone video dataset UAVVIC. Since its test set has not been released, comparisons are performed on the validation set. The results in Table \ref{tab:comparison_on_uavvic} show that our method achieves the best overall performance, demonstrating that our method not only handles dense scenes effectively but also performs well in sparse scenes. UAVVIC contains both static and dynamic drone videos, so we conduct separate tests in both scenarios to ensure a more comprehensive analysis. As shown in Table \ref{tab:comparison_on_uavvic}, the performance of other methods declines significantly in dynamic scenes compared to their performance in static scenes, whereas our method achieves consistently strong results in both settings. This indicates that other methods struggle to handle dynamic scenes with complex motion patterns, while our method performs effectively.

\subsection{Ablation Studies}

\noindent \textbf{Effect of Backbone:}
In our method, image features can be extracted either by CNN or Transformer. Therefore, we first investigate the impact of the backbone. As shown in the first row of Table \ref{tab:ablation_study}, using the VGG-16 backbone yields the best performance. This suggests that CNN can provide richer local details for pixel-level tasks such as counting.

\noindent \textbf{Effect of Depth-wise Cross-Frame Attention:}
To invalidate the effectiveness of our proposed DCFA module, we directly use the global features $\mathbf{F}^g_j$ and $\mathbf{F}^g_{j+\delta}$ to compute the cross-frame attention, which we refer to as Shallow-wise Cross-Frame Attention (SCFA). To ensure a fairer comparison, we adjust the hyperparameters in SCFA to ensure its number of parameters is equal to that of DCFA. The results in Table \ref{tab:ablation_study} show that DCFA achieves superior performance, as it effectively integrates multi-scale features while learning shared pedestrian information across adjacent frames. 

\noindent \textbf{Effect of Position Embedding in DCFA:}
The experimental results in Table \ref{tab:ablation_study} show that positional encoding has distinct effects when using different backbones. Specifically, when using CNN as the backbone, incorporating positional encoding in DCFA leads to a decrease in final performance. However, with a Transformer backbone, adding positional encoding significantly enhances the counting performance. This is because CNN inherently encodes positional information, and adding extra positional encoding may disrupt the semantic integrity of CNN features. In contrast, Transformer features rely on positional encoding to specify the location of each pixel.

\noindent \textbf{Effect of Learning Strategy:}
Our method first predicts the shared density maps, from which the outflow and inflow maps are derived by subtraction from the global density map. To validate the effectiveness of this strategy, we conduct an ablation study where the output of DCFA is decoded and then directly supervised by the ground-truth outflow and inflow density maps, i.e. learning them directly instead of first predicting the shared density map. As shown in the seventh row of Table \ref{tab:ablation_study}, directly learning the inflow density map leads to a significant drop in final performance. This suggests that learning shared information between two frames is easier than learning the private information of each frame, further validating the rationality behind our approach.  
\begin{figure}[t]
	\centering
	\includegraphics[width=0.85\linewidth]{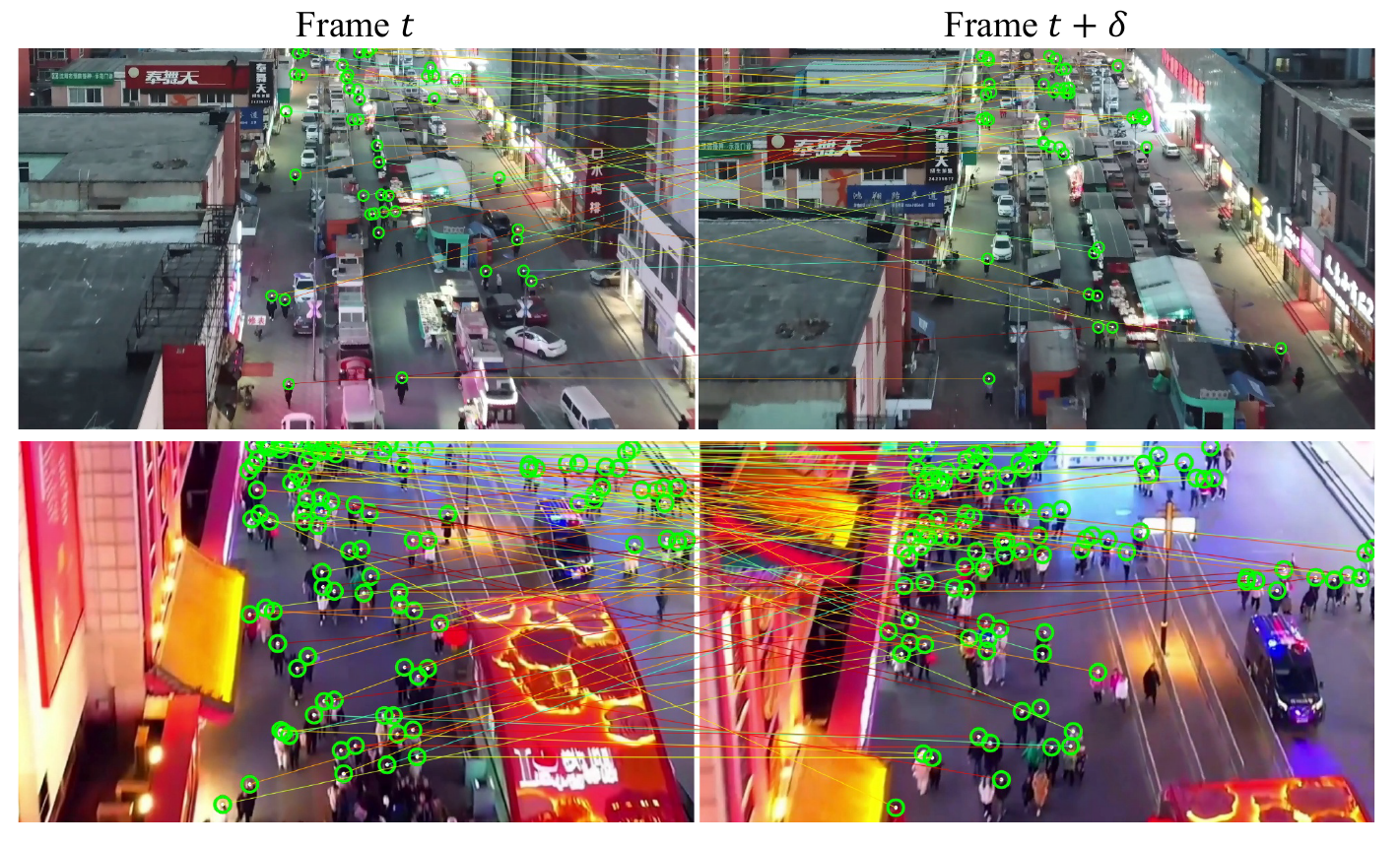}
	\vspace{-0.2cm}
	\caption{The visualization of CGNet on MovingDroneCrowd. There are numerous localization errors in dense scenes, and the cross-frame association are almost entirely incorrect.}
	\label{fig:compare_visual}
    \vspace{-0.8cm}
\end{figure}

\subsection{Qualitative Results}
Fig. \ref{fig:visual} illustrates the visual results of our method on examples of MovingDroneCrowd. The inflow and outflow density maps reflect pedestrian entries and exits within the field of view. Although some erroneous responses exist, their values are effectively suppressed.
Fig. \ref{fig:compare_visual} presents the visual results of CGNet on the same image pairs. Significant errors are observed in both localization and association, with the association being almost entirely incorrect. This suggests that previous localization and association-based methods struggle to handle dynamic and dense scenes effectively.

\section{Conclusion}
This paper explores a flexible approach to counting unique individuals over a large area in a period of time, specifically in videos captured by moving drones. Due to the lack of relevant datasets and effective algorithms, we introduce MovingDroneCrowd, a challenging video-level dataset captured by moving drones in crowded scenes with diverse lighting, altitudes, angles, and complex motion patterns. These factors make previous location-based methods ineffective. Therefore, we propose a density map-based algorithm for video individual counting that bypass localization and association. Instead, we directly estimate the inflow density map, which reflects the number of newly entered crowd. Experiments on both our and previous benchmarks demonstrate that our method effectively handles high-density and dynamic scenes while also achieving excellent results in static and sparse scenarios.

\section*{Acknowledgements}
This work was supported partially by the National Natural Science Foundation of China (U22A2095, 62276281, 62406090) and Guangdong Basic and Applied Basic Research Foundation, China (2024A1515011882).
\clearpage
{
    \small
    \bibliographystyle{ieeenat_fullname}
    \bibliography{main}

\begin{thebibliography}{48}
\providecommand{\natexlab}[1]{#1}
\providecommand{\url}[1]{\texttt{#1}}
\expandafter\ifx\csname urlstyle\endcsname\relax
  \providecommand{\doi}[1]{doi: #1}\else
  \providecommand{\doi}{doi: \begingroup \urlstyle{rm}\Url}\fi

\bibitem[Aharon et~al.(2022)Aharon, Orfaig, and Bobrovsky]{BoT_SORT}
Nir Aharon, Roy Orfaig, and Ben-Zion Bobrovsky.
\newblock Bot-sort: Robust associations multi-pedestrian tracking.
\newblock \emph{arXiv preprint arXiv:2206.14651}, 2022.

\bibitem[Bahmanyar et~al.(2019)Bahmanyar, Vig, and Reinartz]{bahmanyar2019mrcnet}
Reza Bahmanyar, Elenora Vig, and Peter Reinartz.
\newblock Mrcnet: Crowd counting and density map estimation in aerial and ground imagery.
\newblock \emph{arXiv preprint arXiv:1909.12743}, 2019.

\bibitem[Cai et~al.(2022)Cai, Xu, Li, Xiong, Xia, Tu, and Soatto]{MeMOT}
Jiarui Cai, Mingze Xu, Wei Li, Yuanjun Xiong, Wei Xia, Zhuowen Tu, and Stefano Soatto.
\newblock Memot: Multi-object tracking with memory.
\newblock In \emph{Proceedings of the IEEE/CVF Conference on Computer Vision and Pattern Recognition (CVPR)}, pages 8090--8100, 2022.

\bibitem[Cao et~al.(2023)Cao, Pang, Weng, Khirodkar, and Kitani]{OC-SORT}
Jinkun Cao, Jiangmiao Pang, Xinshuo Weng, Rawal Khirodkar, and Kris Kitani.
\newblock Observation-centric sort: Rethinking sort for robust multi-object tracking.
\newblock In \emph{Proceedings of the IEEE/CVF Conference on Computer Vision and Pattern Recognition}, pages 9686--9696, 2023.

\bibitem[Chan and Vasconcelos(2009)]{Bayesian_Poisson}
Antoni~B. Chan and Nuno Vasconcelos.
\newblock Bayesian poisson regression for crowd counting.
\newblock In \emph{2009 IEEE 12th International Conference on Computer Vision}, pages 545--551, 2009.

\bibitem[Dong et~al.(2024)Dong, Zhang, Ma, Xu, Yang, and Wu]{CLRNet}
Li Dong, Haijun Zhang, Jianghong Ma, Xiaofei Xu, Yimin Yang, and Q.~M.~Jonathan Wu.
\newblock Clrnet: A cross locality relation network for crowd counting in videos.
\newblock \emph{IEEE Transactions on Neural Networks and Learning Systems}, 35\penalty0 (5):\penalty0 6408--6422, 2024.

\bibitem[Du et~al.(2023{\natexlab{a}})Du, Deng, and Shi]{Domain-General_Du_Deng_Shi_2023}
Zhipeng Du, Jiankang Deng, and Miaojing Shi.
\newblock Domain-general crowd counting in unseen scenarios.
\newblock \emph{Proceedings of the AAAI Conference on Artificial Intelligence}, 37\penalty0 (1):\penalty0 561--570, 2023{\natexlab{a}}.

\bibitem[Du et~al.(2023{\natexlab{b}})Du, Shi, Deng, and Zafeiriou]{Redesigning_Multi-Scale}
Zhipeng Du, Miaojing Shi, Jiankang Deng, and Stefanos Zafeiriou.
\newblock Redesigning multi-scale neural network for crowd counting.
\newblock \emph{IEEE Transactions on Image Processing}, 32:\penalty0 3664--3678, 2023{\natexlab{b}}.

\bibitem[Fan et~al.(2025)Fan, Wan, and Ma]{LCSD}
Yaowu Fan, Jia Wan, and Andy~J. Ma.
\newblock Learning crowd scale and distribution for weakly supervised crowd counting and localization.
\newblock \emph{IEEE Transactions on Circuits and Systems for Video Technology}, 35\penalty0 (1):\penalty0 713--727, 2025.

\bibitem[Fang et~al.(2019)Fang, Zhan, Cai, Gao, and Hu]{LCSTNVCC}
Yanyan Fang, Biyun Zhan, Wandi Cai, Shenghua Gao, and Bo Hu.
\newblock Locality-constrained spatial transformer network for video crowd counting.
\newblock In \emph{2019 IEEE International Conference on Multimedia and Expo (ICME)}, pages 814--819, 2019.

\bibitem[Gao et~al.(2023)Gao, Han, Yuan, and Wang]{HQITDR}
Junyu Gao, Tao Han, Yuan Yuan, and Qi Wang.
\newblock Domain-adaptive crowd counting via high-quality image translation and density reconstruction.
\newblock \emph{IEEE Transactions on Neural Networks and Learning Systems}, 34\penalty0 (8):\penalty0 4803--4815, 2023.

\bibitem[Guo et~al.(2024)Guo, Yuan, Yan, Chen, Wang, and Ye]{RSMPL}
Mingyue Guo, Li Yuan, Zhaoyi Yan, Binghui Chen, Yaowei Wang, and Qixiang Ye.
\newblock Regressor-segmenter mutual prompt learning for crowd counting.
\newblock In \emph{2024 IEEE/CVF Conference on Computer Vision and Pattern Recognition (CVPR)}, pages 28380--28389, 2024.

\bibitem[Han et~al.(2020)Han, Gao, Yuan, and Wang]{FSCCDCU}
Tao Han, Junyu Gao, Yuan Yuan, and Qi Wang.
\newblock Focus on semantic consistency for cross-domain crowd understanding.
\newblock In \emph{ICASSP 2020 - 2020 IEEE International Conference on Acoustics, Speech and Signal Processing (ICASSP)}, pages 1848--1852, 2020.

\bibitem[Han et~al.(2022)Han, Bai, Gao, Wang, and Ouyang]{DRNet}
Tao Han, Lei Bai, Junyu Gao, Qi Wang, and Wanli Ouyang.
\newblock Dr.vic: Decomposition and reasoning for video individual counting.
\newblock In \emph{Proceedings of the IEEE/CVF Conference on Computer Vision and Pattern Recognition (CVPR)}, pages 3083--3092, 2022.

\bibitem[Han et~al.(2023)Han, Bai, Liu, and Ouyang]{STEERER}
Tao Han, Lei Bai, Lingbo Liu, and Wanli Ouyang.
\newblock Steerer: Resolving scale variations for counting and localization via selective inheritance learning.
\newblock In \emph{Proceedings of the IEEE/CVF International Conference on Computer Vision (ICCV)}, pages 21848--21859, 2023.

\bibitem[Idrees et~al.(2013)Idrees, Saleemi, Seibert, and Shah]{MSMS}
Haroon Idrees, Imran Saleemi, Cody Seibert, and Mubarak Shah.
\newblock Multi-source multi-scale counting in extremely dense crowd images.
\newblock In \emph{Proceedings of the IEEE Conference on Computer Vision and Pattern Recognition (CVPR)}, 2013.

\bibitem[Jiang et~al.(2020)Jiang, Zhang, Xu, Zhang, Lv, Zhou, Yang, and Pang]{Attention_Scaling}
Xiaoheng Jiang, Li Zhang, Mingliang Xu, Tianzhu Zhang, Pei Lv, Bing Zhou, Xin Yang, and Yanwei Pang.
\newblock Attention scaling for crowd counting.
\newblock In \emph{Proceedings of the IEEE/CVF Conference on Computer Vision and Pattern Recognition (CVPR)}, 2020.

\bibitem[Lempitsky and Zisserman(2010)]{LTCO}
Victor Lempitsky and Andrew Zisserman.
\newblock Learning to count objects in images.
\newblock In \emph{Advances in Neural Information Processing Systems}. Curran Associates, Inc., 2010.

\bibitem[Li et~al.(2022)Li, Liu, Yang, Liu, Gao, Zhao, Zhang, and Hou]{VSCrowd}
Haopeng Li, Lingbo Liu, Kunlin Yang, Shinan Liu, Junyu Gao, Bin Zhao, Rui Zhang, and Jun Hou.
\newblock Video crowd localization with multifocus gaussian neighborhood attention and a large-scale benchmark.
\newblock \emph{IEEE Transactions on Image Processing}, 31:\penalty0 6032--6047, 2022.

\bibitem[Li et~al.(2024)Li, Liu, Li, Li, and Lu]{PGDTR}
Rui Li, Yishu Liu, Huafeng Li, Jinxing Li, and Guangming Lu.
\newblock Prototype-guided dual-transformer reasoning for video individual counting.
\newblock page 10258–10267, 2024.

\bibitem[Li et~al.(2018)Li, Zhang, and Chen]{csrnet}
Yuhong Li, Xiaofan Zhang, and Deming Chen.
\newblock Csrnet: Dilated convolutional neural networks for understanding the highly congested scenes.
\newblock In \emph{Proceedings of the IEEE Conference on Computer Vision and Pattern Recognition (CVPR)}, 2018.

\bibitem[Liang et~al.(2022)Liang, Xu, and Bai]{E2ETCL}
Dingkang Liang, Wei Xu, and Xiang Bai.
\newblock An end-to-end transformer model for crowd localization.
\newblock In \emph{Proceedings of the European Conference on Computer Vision (ECCV)}, pages 38--54, 2022.

\bibitem[Liu et~al.(2019)Liu, Salzmann, and Fua]{Liu_2019_CVPR_Context_Aware}
Weizhe Liu, Mathieu Salzmann, and Pascal Fua.
\newblock Context-aware crowd counting.
\newblock In \emph{Proceedings of the IEEE/CVF Conference on Computer Vision and Pattern Recognition (CVPR)}, 2019.

\bibitem[Liu et~al.(2022)Liu, Durasov, and Fua]{liu2022leveraging}
Weizhe Liu, Nikita Durasov, and Pascal Fua.
\newblock Leveraging self-supervision for cross-domain crowd counting.
\newblock In \emph{Proceedings of the IEEE/CVF Conference on Computer Vision and Pattern Recognition}, pages 5341--5352, 2022.

\bibitem[Liu et~al.(2024)Liu, Li, Qi, Yan, Han, van~den Hengel, Yang, and Huang]{uavvic}
Xinyan Liu, Guorong Li, Yuankai Qi, Ziheng Yan, Zhenjun Han, Anton van~den Hengel, Ming-Hsuan Yang, and Qingming Huang.
\newblock Weakly supervised video individual counting.
\newblock In \emph{Proceedings of the IEEE/CVF Conference on Computer Vision and Pattern Recognition (CVPR)}, pages 19228--19237, 2024.

\bibitem[Liu et~al.(2021)Liu, He, Wang, Wang, Yuan, Zhang, Zhang, Zhu, Gool, Han, Hoi, Hu, Liu, Pan, Yin, Zhang, Liu, Ding, Liang, Ding, Lu, Lin, Chen, Li, Liu, Zhou, Shi, Yang, He, Peng, Xu, Han, Bai, Chen, Wang, Xia, Tao, Chen, and Cao]{VisDrone-CC2021}
Zhihao Liu, Zhijian He, Lujia Wang, Wenguan Wang, Yixuan Yuan, Dingwen Zhang, Jinglin Zhang, Pengfei Zhu, Luc~Van Gool, Junwei Han, Steven Hoi, Qinghua Hu, Ming Liu, Junwen Pan, Baoqun Yin, Binyu Zhang, Chengxin Liu, Ding Ding, Dingkang Liang, Guanchen Ding, Hao Lu, Hui Lin, Jingyuan Chen, Jiong Li, Liang Liu, Lin Zhou, Min Shi, Qianqian Yang, Qing He, Sifan Peng, Wei Xu, Wenwei Han, Xiang Bai, Xiwu Chen, Yabin Wang, Yinfeng Xia, Yiran Tao, Zhenzhong Chen, and Zhiguo Cao.
\newblock Visdrone-cc2021: The vision meets drone crowd counting challenge results.
\newblock In \emph{2021 IEEE/CVF International Conference on Computer Vision Workshops (ICCVW)}, pages 2830--2838, 2021.

\bibitem[Lowe(1999)]{OCLSI}
D.G. Lowe.
\newblock Object recognition from local scale-invariant features.
\newblock In \emph{Proceedings of the Seventh IEEE International Conference on Computer Vision}, pages 1150--1157, 1999.

\bibitem[Lv et~al.(2024)Lv, Huang, Zhang, Lin, Han, and Zeng]{DiffMOT}
Weiyi Lv, Yuhang Huang, Ning Zhang, Ruei-Sung Lin, Mei Han, and Dan Zeng.
\newblock Diffmot: A real-time diffusion-based multiple object tracker with non-linear prediction.
\newblock In \emph{Proceedings of the IEEE/CVF Conference on Computer Vision and Pattern Recognition (CVPR)}, pages 19321--19330, 2024.

\bibitem[Meinhardt et~al.(2022)Meinhardt, Kirillov, Leal-Taix\'e, and Feichtenhofer]{TrackFormer}
Tim Meinhardt, Alexander Kirillov, Laura Leal-Taix\'e, and Christoph Feichtenhofer.
\newblock Trackformer: Multi-object tracking with transformers.
\newblock In \emph{Proceedings of the IEEE/CVF Conference on Computer Vision and Pattern Recognition (CVPR)}, pages 8844--8854, 2022.

\bibitem[Peng et~al.(2021)Peng, Li, and Zhu]{RGB-T_CCD}
Tao Peng, Qing Li, and Pengfei Zhu.
\newblock Rgb-t crowd counting from drone: A benchmark and mmccn network.
\newblock In \emph{Computer Vision -- ACCV 2020}, pages 497--513, 2021.

\bibitem[Shi et~al.(2019)Shi, Yang, Xu, and Chen]{Reverse_Perspective2}
Miaojing Shi, Zhaohui Yang, Chao Xu, and Qijun Chen.
\newblock Revisiting perspective information for efficient crowd counting.
\newblock In \emph{Proceedings of the IEEE/CVF Conference on Computer Vision and Pattern Recognition (CVPR)}, 2019.

\bibitem[Song et~al.(2021)Song, Wang, Jiang, Wang, Tai, Wang, Li, Huang, and Wu]{P2P}
Qingyu Song, Changan Wang, Zhengkai Jiang, Yabiao Wang, Ying Tai, Chengjie Wang, Jilin Li, Feiyue Huang, and Yang Wu.
\newblock Rethinking counting and localization in crowds: A purely point-based framework.
\newblock In \emph{Proceedings of the IEEE/CVF International Conference on Computer Vision (ICCV)}, pages 3365--3374, 2021.

\bibitem[Sun et~al.(2022)Sun, Cao, Jiang, Yuan, Bai, Kitani, and Luo]{DanceTrack}
Peize Sun, Jinkun Cao, Yi Jiang, Zehuan Yuan, Song Bai, Kris Kitani, and Ping Luo.
\newblock Dancetrack: Multi-object tracking in uniform appearance and diverse motion.
\newblock In \emph{Proceedings of the IEEE/CVF Conference on Computer Vision and Pattern Recognition (CVPR)}, pages 20993--21002, 2022.

\bibitem[Sundararaman et~al.(2021)Sundararaman, De~Almeida~Braga, Marchand, and Pettre]{Headhunter-T}
Ramana Sundararaman, Cedric De~Almeida~Braga, Eric Marchand, and Julien Pettre.
\newblock Tracking pedestrian heads in dense crowd.
\newblock In \emph{Proceedings of the IEEE/CVF Conference on Computer Vision and Pattern Recognition (CVPR)}, pages 3865--3875, 2021.

\bibitem[Wan et~al.(2024)Wan, Huang, and Shuai]{FMDC}
Chang-Lin Wan, Feng-Kai Huang, and Hong-Han Shuai.
\newblock Density-based flow mask integration via deformable convolution for video people flux estimation.
\newblock In \emph{Proceedings of the IEEE/CVF Winter Conference on Applications of Computer Vision (WACV)}, pages 6573--6582, 2024.

\bibitem[Wang et~al.(2023)Wang, Cai, Han, Zhou, and Gong]{STNet}
Mingjie Wang, Hao Cai, Xian-Feng Han, Jun Zhou, and Minglun Gong.
\newblock Stnet: Scale tree network with multi-level auxiliator for crowd counting.
\newblock \emph{IEEE Transactions on Multimedia}, 25:\penalty0 2074--2084, 2023.

\bibitem[Wang et~al.(2022)Wang, Han, Gao, and Yuan]{NLT}
Qi Wang, Tao Han, Junyu Gao, and Yuan Yuan.
\newblock Neuron linear transformation: Modeling the domain shift for crowd counting.
\newblock \emph{IEEE Transactions on Neural Networks and Learning Systems}, 33\penalty0 (8):\penalty0 3238--3250, 2022.

\bibitem[Wang et~al.(2020)Wang, Zheng, Liu, Li, and Wang]{TRMOT}
Zhongdao Wang, Liang Zheng, Yixuan Liu, Yali Li, and Shengjin Wang.
\newblock Towards real-time multi-object tracking.
\newblock In \emph{Computer Vision -- ECCV 2020}, pages 107--122, 2020.

\bibitem[Wen et~al.(2019)Wen, Du, Zhu, Hu, Wang, Bo, and Lyu]{DBJDLT}
Longyin Wen, Dawei Du, Pengfei Zhu, Qinghua Hu, Qilong Wang, Liefeng Bo, and Siwei Lyu.
\newblock Drone-based joint density map estimation, localization and tracking with space-time multi-scale attention network.
\newblock \emph{arXiv preprint arXiv:1912.01811}, 2019.

\bibitem[Wen et~al.(2021)Wen, Du, Zhu, Hu, Wang, Bo, and Lyu]{DTCMDC}
Longyin Wen, Dawei Du, Pengfei Zhu, Qinghua Hu, Qilong Wang, Liefeng Bo, and Siwei Lyu.
\newblock Detection, tracking, and counting meets drones in crowds: A benchmark.
\newblock In \emph{Proceedings of the IEEE/CVF Conference on Computer Vision and Pattern Recognition (CVPR)}, pages 7812--7821, 2021.

\bibitem[Xie et~al.(2023)Xie, Yang, Zhu, and Wang]{Striking_a_Balance}
Haiyang Xie, Zhengwei Yang, Huilin Zhu, and Zheng Wang.
\newblock Striking a balance: Unsupervised cross-domain crowd counting via knowledge diffusion.
\newblock In \emph{Proceedings of the 31st ACM International Conference on Multimedia}, page 6520–6529, 2023.

\bibitem[Yan et~al.(2019)Yan, Yuan, Zuo, Tan, Wang, Wen, and Ding]{Perspective-Guided}
Zhaoyi Yan, Yuchen Yuan, Wangmeng Zuo, Xiao Tan, Yezhen Wang, Shilei Wen, and Errui Ding.
\newblock Perspective-guided convolution networks for crowd counting.
\newblock In \emph{Proceedings of the IEEE/CVF International Conference on Computer Vision (ICCV)}, 2019.

\bibitem[Yang et~al.(2020)Yang, Li, Wu, Su, Huang, and Sebe]{Reverse_Perspective}
Yifan Yang, Guorong Li, Zhe Wu, Li Su, Qingming Huang, and Nicu Sebe.
\newblock Reverse perspective network for perspective-aware object counting.
\newblock In \emph{Proceedings of the IEEE/CVF Conference on Computer Vision and Pattern Recognition (CVPR)}, 2020.

\bibitem[Zhan et~al.(2008)Zhan, Monekosso, Remagnino, Velastin, and Xu]{Crowd_analysis_survey}
Biao Zhan, Dorothy~N Monekosso, Paolo Remagnino, Sergio~A Velastin, and Li-Qun Xu.
\newblock Crowd analysis: a survey.
\newblock \emph{Machine Vision and Applications}, 19\penalty0 (5):\penalty0 345--357, 2008.

\bibitem[Zhang et~al.(2016)Zhang, Zhou, Chen, Gao, and Ma]{Zhang_2016_CVPR_MCNN}
Yingying Zhang, Desen Zhou, Siqin Chen, Shenghua Gao, and Yi Ma.
\newblock Single-image crowd counting via multi-column convolutional neural network.
\newblock In \emph{Proceedings of the IEEE Conference on Computer Vision and Pattern Recognition (CVPR)}, 2016.

\bibitem[Zhang et~al.(2022)Zhang, Sun, Jiang, Yu, Weng, Yuan, Luo, Liu, and Wang]{ByteTrack}
Yifu Zhang, Peize Sun, Yi Jiang, Dongdong Yu, Fucheng Weng, Zehuan Yuan, Ping Luo, Wenyu Liu, and Xinggang Wang.
\newblock Bytetrack: Multi-object tracking by associating every detection box.
\newblock In \emph{Computer Vision -- ECCV 2022}, pages 1--21, 2022.

\bibitem[Zhao et~al.(2016)Zhao, Li, Zhao, and Wang]{LOI}
Zhuoyi Zhao, Hongsheng Li, Rui Zhao, and Xiaogang Wang.
\newblock Crossing-line crowd counting with two-phase deep neural networks.
\newblock In \emph{Computer Vision -- ECCV 2016}, pages 712--726, 2016.

\bibitem[Zhu et~al.(2021)Zhu, Wen, Du, Bian, Fan, Hu, and Ling]{VisDroneDatasets}
Pengfei Zhu, Longyin Wen, Dawei Du, Xiao Bian, Heng Fan, Qinghua Hu, and Haibin Ling.
\newblock Detection and tracking meet drones challenge.
\newblock \emph{IEEE Transactions on Pattern Analysis and Machine Intelligence}, 44\penalty0 (11):\penalty0 7380--7399, 2021.

\end{thebibliography}
}
\setcounter{page}{1}
\maketitlesupplementary

\appendix

\begin{figure*}[ht]
	\centering
	\includegraphics[width=1\linewidth]{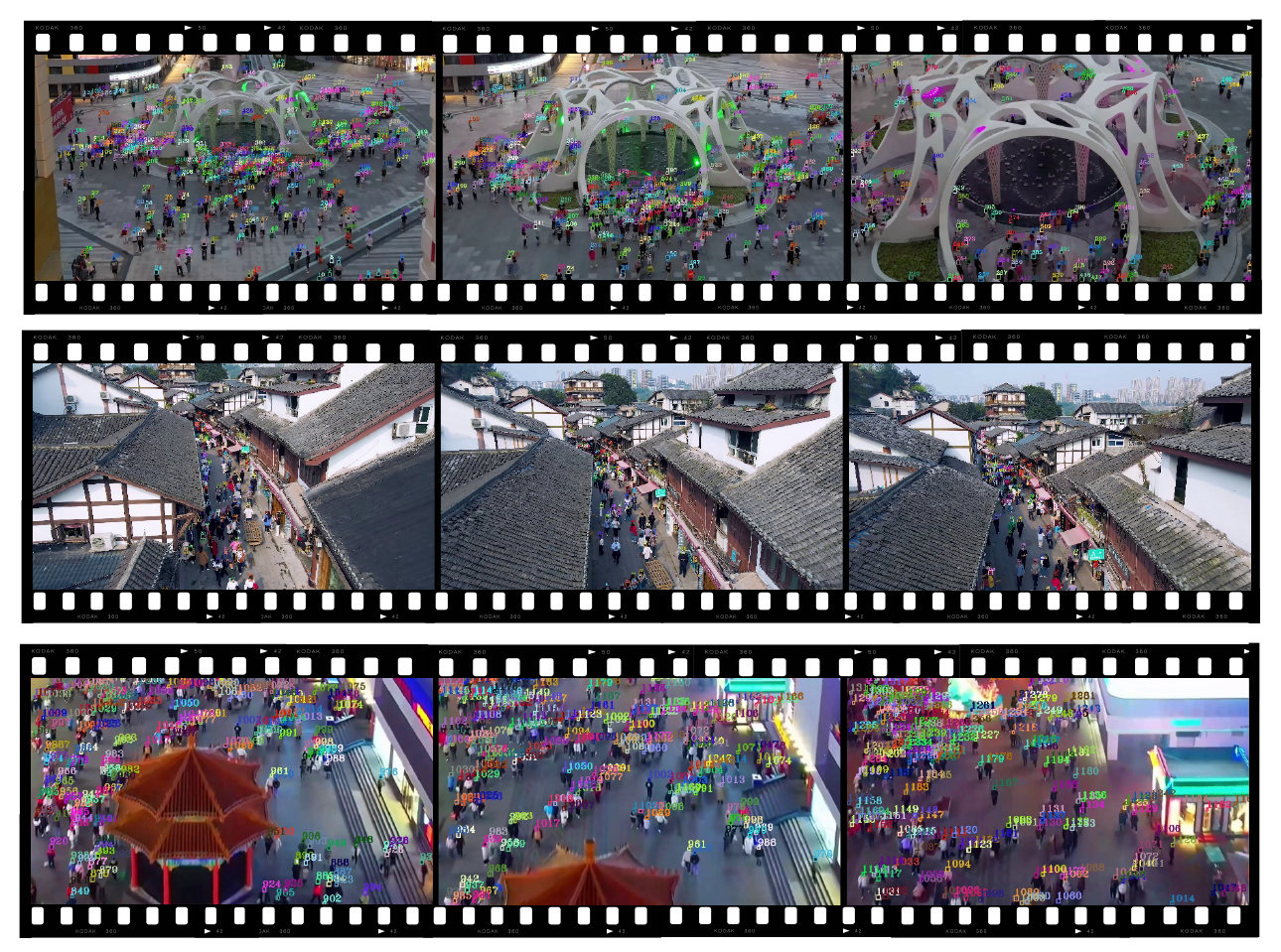}
	\caption{Additional samples from dataset MovingDroneCrowd. Due to space constraints, only three frames from each video are shown, with each frame annotated with head bounding boxes and ID labels.}
	\label{fig:supp_our_data_examples}
    \vspace{-0.5cm}
\end{figure*}
\begin{figure*}[ht]
	\centering
	\includegraphics[width=1\linewidth]{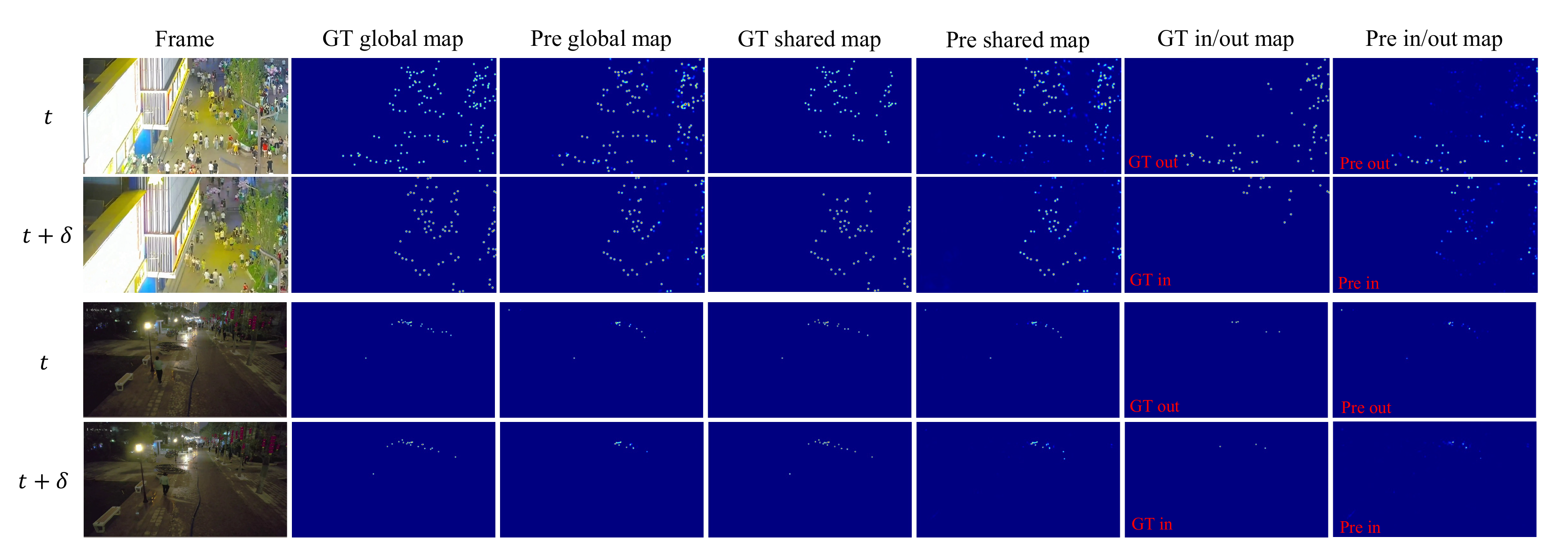}
	\caption{Additional visualization results of our method on dataset MovingDroneCrowd. These results demonstrate that our method performs well in low-light, dense, and sparse scenes.}
	\label{fig:supp_our_visual}
    \vspace{-0.5cm}
\end{figure*}

\begin{figure*}[t]
	\centering
	\includegraphics[width=1\linewidth]{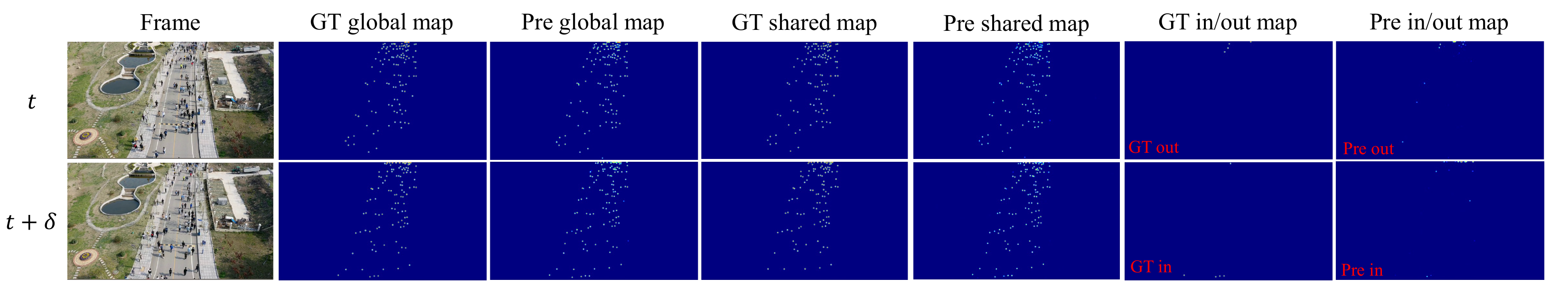}
	\caption{Additional visualization results of our method on dataset UAVVIC. It indicates that our method also achieves satisfactory performance in static scenes.}
	\label{fig:supp_uavvic_visual}
    \vspace{-0.5cm}
\end{figure*}
The supplementary provides more details for the paper ``Video Individual Counting for Moving Drones'', including the following aspects.
 \begin{itemize}[labelindent=2em]
 \item[$\bullet$] Details about Training and Testing.
\item[$\bullet$] More Visualization Results.
\item[$\bullet$] More Examples of MovingDroneCrowd.
\item[$\bullet$] Limitations.
\end{itemize}

\section{Details about Training and Testing}
\textbf{Training details:} Since MovingDroneCrowd videos have been sufficiently  downsampled to eliminate redundancy, we randomly select frame interval $\delta$ in the range of 3 $\sim$ 8 to guarantee the training pairs contain diverse inflow and outflow pedestrian variations. For data augmentation, training images are downsampled so that the longer side does not exceed 2560 pixels and the shorter side does not exceed 1440 pixels, ensuring that the cropped images contain enough pedestrians. The cropping, flipping, and scaling strategies follow those in \cite{DRNet}. The initial learning rate is set as 1e-5 with a weight decay of 1e-6 and follows a polynomial decay with a power of 0.9. We use VGG16, initialized with ImageNet pre-trained weights, as the backbone for feature extraction. The model is implemented with PyTorch and trained on A800 GPUs.

\noindent \textbf{Test details:} Our model can receive images with irregular resolutions during testing. To reduce computational cost, the longer side and shorter sides of the input image are limited to no more than 1920 and 1080 pixels, respectively. 

\begin{figure}[H]
    \centering
        \centering
        \includegraphics[width=\linewidth]{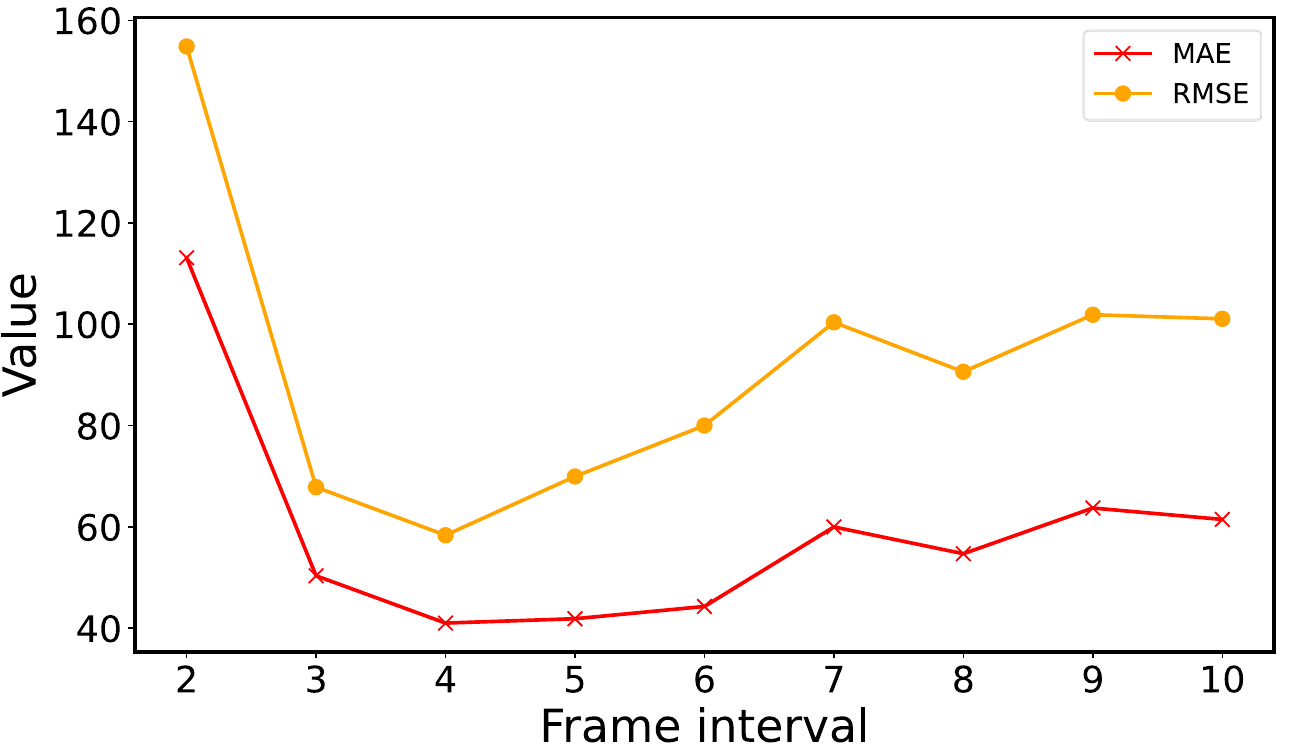}
        \caption{Ablation study of test frame interval $\delta$ on MovingDroneCrowd.}
        \label{fig:test_frame_interval}
\end{figure}

Fig. \ref{fig:test_frame_interval} shows that our method maintains reasonable performance across a wide range of frame intervals, demonstrating its robustness to interval variations. It achieves the best performance when $\delta = 4$, so we set the frame interval $\delta $ to $4$ during testing on MovingDroneCrowd.

\section{More Examples of MovingDroneCrowd}
Fig. \ref{fig:supp_our_data_examples} presents additional video samples from our dataset MovingDroneCrowd, with each frame annotated with head bounding boxes and identity IDs. These examples highlight the key characteristics of our dataset: dense crowds, complex motion patterns, varying lighting conditions, and diverse camera heights and angles.

\section{More Visualization Results}
Fig. \ref{fig:supp_our_visual} presents additional visualization results of our method on 
MovingDroneCrowd. The first scene is a densely crowded scene with significant drone movement, while the second scene captures a sparsely populated area during low-altitude drone flight. Both scenes were recorded under low-light conditions. These results demonstrate that our method accurately predicts the inflow density map for each frame relative to its previous frame. This demonstrates that our method is sufficiently robust, achieving strong performance in complex environments, including dense, sparse, and low-light conditions.

Fig. \ref{fig:supp_uavvic_visual} presents the visualization results of our method on the previous dataset UAVVIC. This scene was captured by a hovering drone with minimal camera movement, demonstrating that our method still performs well in static scenes.

\section{Limitations}
The visualization results on the test set show that the shared density map is not perfectly learned and still contains many erroneous responses, leading to some errors in the inflow and outflow density maps as well. Due to the similarity in pedestrian appearance, directly learning shared pedestrian features across two frames remains a challenging task. Computing cross attention between two frames is computationally expensive and time-consuming. These issues will be addressed in our future work.

\end{document}